\RequirePackage{fix-cm}
\documentclass[smallcondensed]{svjour3}     
\smartqed  
\usepackage{natbib}

\usepackage{times}
\usepackage{epsfig}
\usepackage{graphicx}
\usepackage{amsmath,mathtools}
\usepackage{amssymb}
\usepackage{multirow}
\usepackage{url}
\usepackage{amssymb}
\usepackage{pifont}
\usepackage{array}
\usepackage{graphicx}
\usepackage{subcaption,adjustbox}
\usepackage{soul}
\usepackage{color}
\usepackage[dvipsnames]{xcolor}
\usepackage{breqn}
\usepackage{bm}
\usepackage{hyperref}
\hypersetup{colorlinks,citecolor=blue,linkcolor=blue}
\usepackage{scrtime}
\usepackage[dont-mess-around]{fnpct}
\usepackage{etoolbox}
\newcommand*{\affmark}[1][*]{\textsuperscript{$\dagger$}}
\newcommand{\review}[1]{\textcolor{black}{#1}}

\usepackage[most]{tcolorbox}
\newtcolorbox{highlighted}{colback=yellow,coltext=red,breakable}
\journalname{Springer Artificial Intelligence Review}
\begin{document}

\title{Deep Semantic Segmentation of Natural and Medical Images: A Review
}

\titlerunning{Deep Semantic Segmentation of Natural and Medical Images: A Review}        

\author{Saeid Asgari Taghanaki\affmark[1]\thanks{\textsuperscript{$\dagger$}: Joint first authors}         \and
        Kumar Abhishek\affmark[1]         \and
        Joseph Paul Cohen     \and
        Julien Cohen-Adad        \and
        Ghassan Hamarneh
}

\authorrunning{S. A. Taghanaki et al.} 

\institute{Saeid Asgari Taghanaki, Kumar Abhishek, and Ghassan Hamarneh \at
              School of Computing Science, Simon Fraser University, Canada \\
              \email{\{sasgarit,kabhishe,hamarneh\}@sfu.ca}           
           \and
           Joseph Paul Cohen \at
              Mila, Universit\'{e} de Montr\'{e}al, Canada\\
              \email{joseph@josephpcohen.com}
                         \and
           Julien Cohen-Adad \at
              NeuroPoly Lab, Institute of Biomedical Engineering, Polytechnique Montr\'{e}al, Canada\\
              \email{jcohen@polymtl.ca}
}

\date{Received: 20 December 2019 / Accepted: 25 May 2020}

\maketitle

\begin{abstract}
The semantic image segmentation task consists of classifying each pixel of an image into an instance, where each instance corresponds to a class. This task is a part of the concept of scene understanding or better explaining the global context of an image. In the medical image analysis domain, image segmentation can be used for image-guided interventions, radiotherapy, or improved radiological diagnostics. In this review, we categorize the leading deep learning-based medical and non-medical image segmentation solutions into six main groups of deep architectural, data synthesis-based, loss function-based, sequenced models, weakly supervised, and multi-task methods~\review{ and provide a comprehensive review of the contributions in each of these groups}. Further, for each group, we analyze each variant of these groups and discuss the limitations of the current approaches and present potential future research directions for semantic image segmentation.
\keywords{Semantic Image Segmentation \and Deep Learning}
\end{abstract}

\section{Introduction}

Deep learning has had a tremendous impact on various fields in science. The focus of the current study is on one of the most critical areas of computer vision: medical image analysis (or medical computer vision), particularly deep learning-based approaches for medical image segmentation. Segmentation is an important processing step in natural images for scene understanding and medical image analysis, for image-guided interventions, radiotherapy, or improved radiological diagnostics, etc. \review{Image segmentation is formally defined as ``the partition of an image into a set of nonoverlapping
regions whose union is the entire image''~\citep{haralick_shapiro_1992}}. A plethora of deep learning approaches for medical image segmentation have been introduced in the literature for different medical imaging modalities, including X-ray, visible-light imaging (e.g. colour dermoscopic images), magnetic resonance imaging (MRI), positron emission tomography (PET), computerized tomography (CT), and ultrasound (e.g. echocardiographic scans). Deep architectural improvement has been a focus of many researchers for different purposes, e.g., tackling gradient vanishing and exploding of deep models, model compression for efficient small yet accurate models, while other works have tried to improve the performance of deep networks by introducing new optimization functions. 

\review{\cite{guo2018review} provided a review of deep learning based semantic segmentation of images, and divided the literature into three categories: region-based, fully convolutional network (FCN)-based, and weakly supervised segmentation methods. \cite{hu2018rgb} summarized the most commonly used RGB-D datasets for semantic segmentation as well as traditional machine learning based methods and deep learning-based network architectures for RGB-D segmentation. \cite{lateef2019survey} presented an extensive survey of deep learning architectures, datasets, and evaluation methods for the semantic segmentation of natural images using deep neural networks. Similarly, for medical imaging, \cite{goceri2017deep} presented an high-level overview of deep learning-based medical image analysis techniques and application areas. \cite{hesamian2019deep} presented an overview of the state-of-the-art methods in medical image segmentation using deep learning by covering the literature related to network structures and model training techniques. \cite{karimi2019deep} reviewed the literature on techniques to handle label noise in deep learning based medical image analysis and evaluated existing approaches on three medical imaging datasets for segmentation and classification tasks. \cite{zhou2019review} presented a review of techniques proposed for fusion of medical images from multiple modalities for medical image segmentation. \cite{goceri2019challenges} discussed the fully supervised, weakly supervised and transfer learning techniques for training deep neural networks for segmentation of medical images, and also discussed the existing methods for addressing the problems of lack of data and class imbalance. \cite{zhang2019survey} presented a review of the approaches to address the problem of small sample sizes in medical image analysis, and divided the literature into five categories including explanation, weakly supervised, transfer learning, and active learning techniques. \cite{tajbakhsh2020embracing} presented a review of the literature for addressing the challenges of scarce annotations as well as weak annotations (e.g., noisy annotations, image-level labels, sparse annotations, etc.) in medical image segmentation. Similarly, there are several surveys covering the literature on the task of object detection~\citep{wang2019salient,zou2019object,Borji2019,Liu2019,Zhao2019}, which can also be used to obtain what can be termed as rough localizations of the object(s) of interest. In contrast to the existing surveys, we make the following contributions in this review:}

\begin{figure}[!ht]
    \centering
    \begin{subfigure}[t]{\textwidth}
        \centering
        {\includegraphics[width=\textwidth]{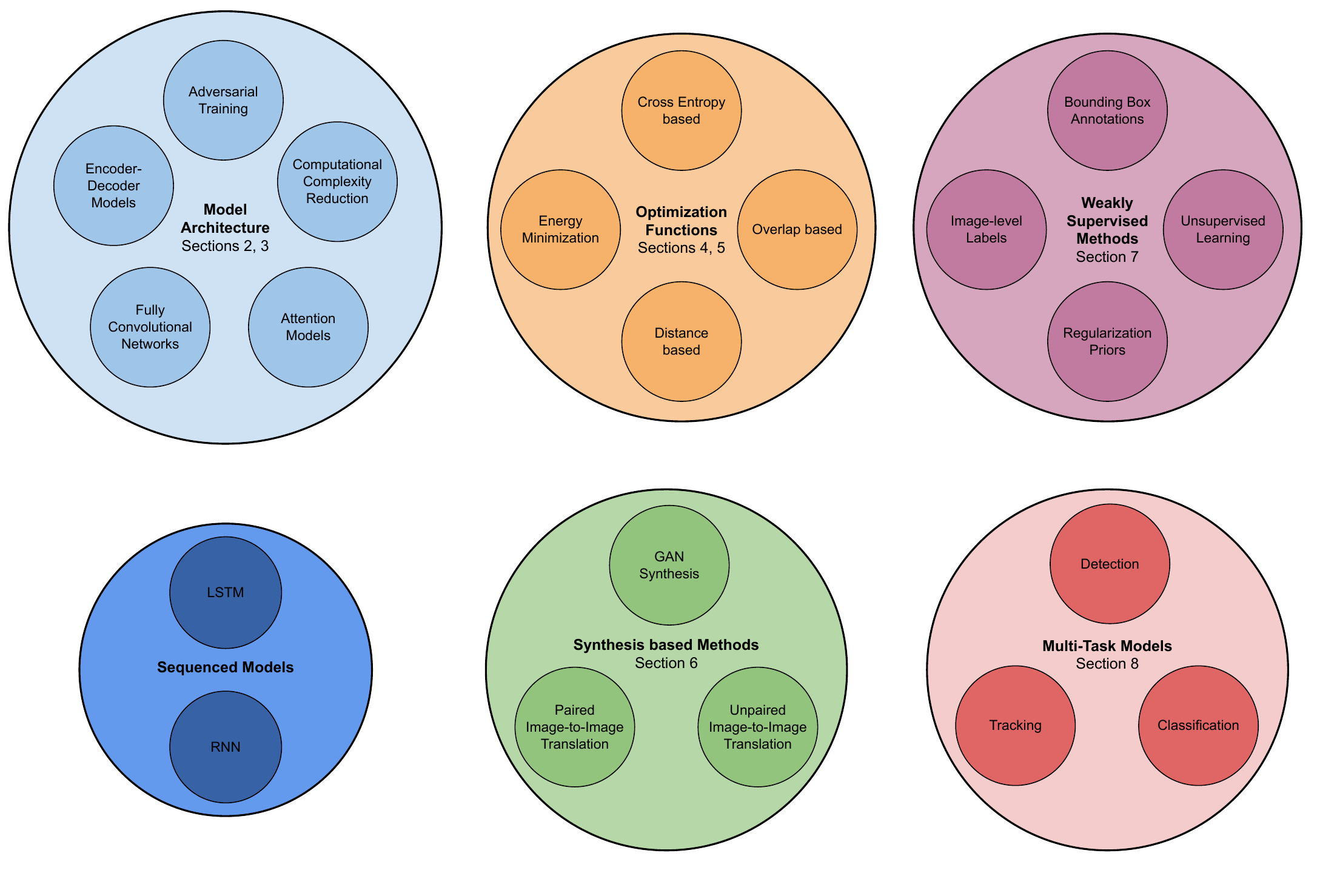}}
        \caption{Topics surveyed in this review.}
    \end{subfigure}\\
    
    \begin{subfigure}[t]{\textwidth}
        \centering
        {\includegraphics[width=\textwidth]{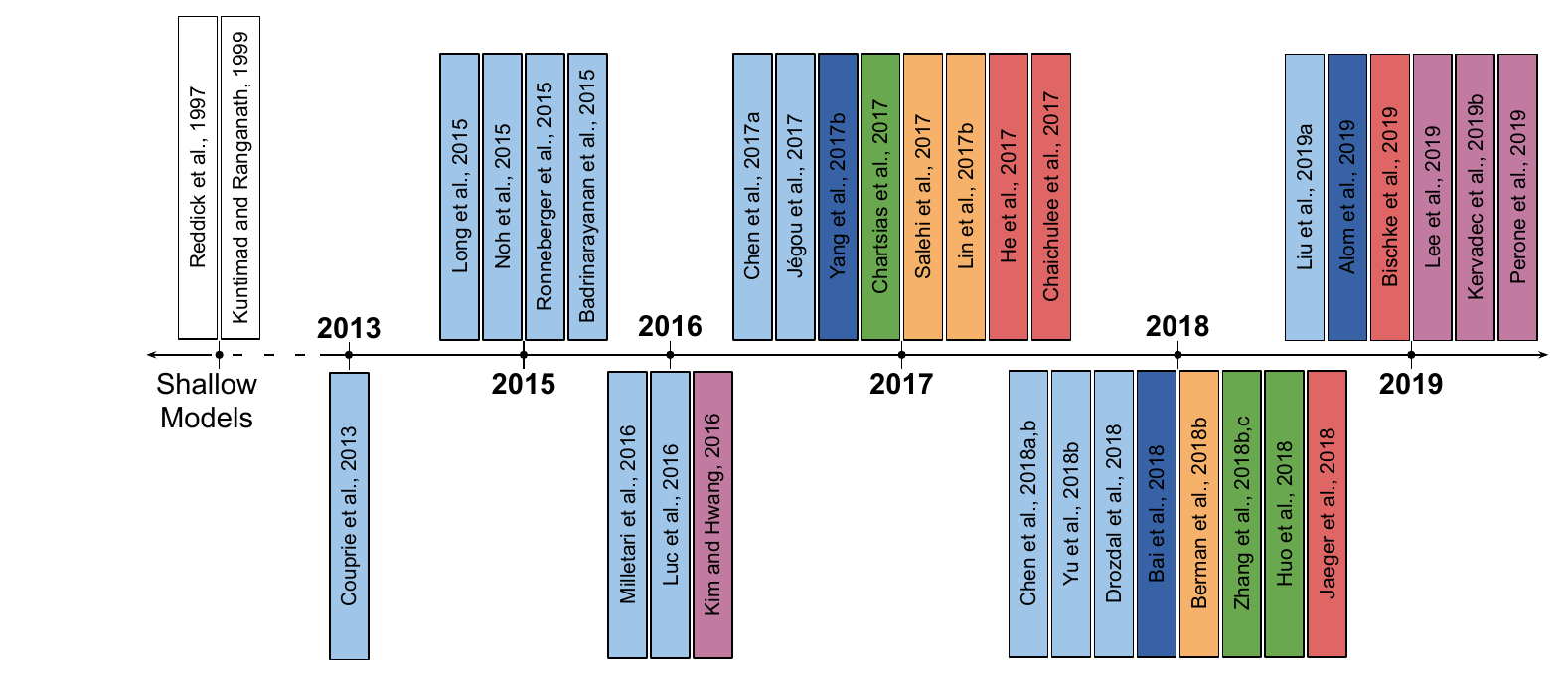}}
        \caption{\review{A timeline of the various contributions in deep learning based semantic segmentation of natural and medical images. The contributions are colored according to their topics in (a) above.}}
    \end{subfigure}\\
    \caption{\review{An overview of the deep learning based segmentation methods covered in this review.}}
    \label{fig:review}
\end{figure}
\begin{figure}[!ht]
\centering
\includegraphics[width=\textwidth]{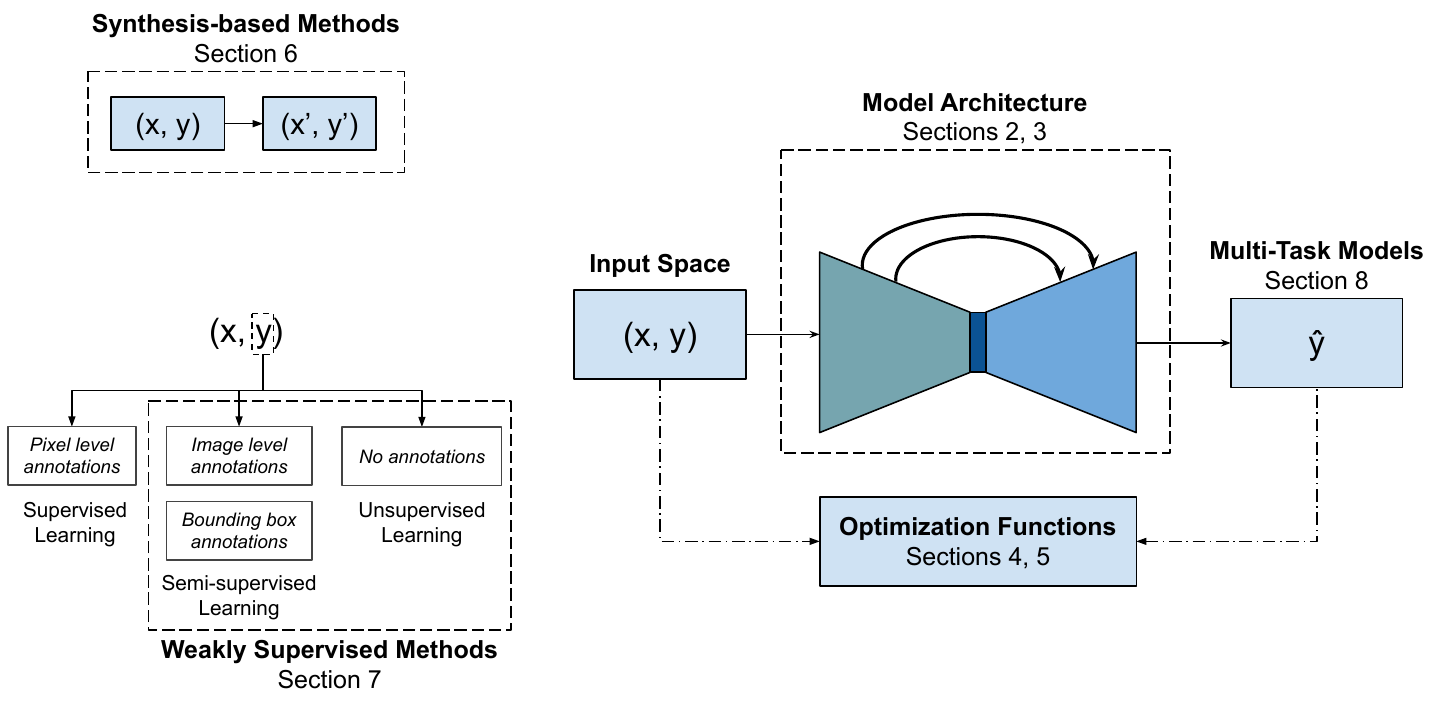}
\caption{\review{A typical deep neural network based semantic segmentation pipeline. Each component in the pipeline indicates the section of this paper that covers the corresponding contributions.}}
\label{fig:overview}
\end{figure}

\begin{itemize}
    \item We provide comprehensive coverage of research contributions in the field of semantic segmentation of natural and medical images. In terms of medical imaging modalities, we cover \review{the literature pertaining to} both 2D (RGB and grayscale) as well as volumetric medical images.
    \item We group the semantic image segmentation literature into six different categories based on the nature of their contributions: architectural improvements, optimization function based improvements, data synthesis based improvements, weakly supervised models, sequenced models, and multi-task models. Figure~\ref{fig:review} indicates the categories we cover in this review\review{, along with a timeline of the most influential papers in the respective categories. Moreover, Figure~\ref{fig:overview} shows a high-level overview of the deep semantic segmentation pipeline, and where each of the categories mentioned in Figure~\ref{fig:review} belong in the pipeline.}
    \item \review{We study the behaviour of many popular loss functions used to train segmentation models on handling scenarios with varying levels of false positive and negative predictions.}
    \item Followed by the comprehensive review, we recognize and suggest the important research directions for each of the categories.
\end{itemize}


In the following sections, we discuss deep semantic image segmentation improvements under different categories visualized in Figure~\ref{fig:review}. For each category, we first review the improvements on non-medical datasets, and in a subsequent section, we survey the improvements for medical images.

\section{Network Architectural Improvements}
\label{netimprove}
This section discusses the advancements in semantic image segmentation using convolutional neural networks (CNNs), which have been applied to interpretation tasks on both natural and medical images~\citep{garcia2018survey,litjens2017survey}. \review{Although artificial neural network-based image segmentation approaches have been explored in the past using shallow networks~\citep{Reddick1997,Kuntimad1999} as well as works which relied on superpixel segmentation maps to generate pixelwise predictions~\citep{couprie2013indoor}, in this work, we focus on deep neural network based image segmentation models which are end-to-end trainable.} The improvements are mostly attributed to exploring new neural architectures (with varying depths, widths, and connectivity or topology) or designing new types of components or layers. 

\subsection{Fully Convolutional Neural Networks for Semantic Segmentation}

\begin{figure}[!ht]
\centering
\includegraphics[width=.7\textwidth]{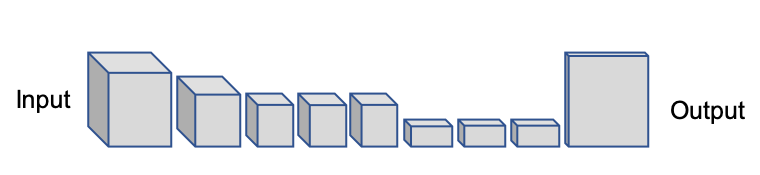}
\caption {Fully convolutional networks can efficiently learn to make dense predictions for per-pixel tasks like semantic segmentation~\citep{long2015fully}.}
\label{fully_conv}
\end{figure}

As one of the first high impact CNN-based segmentation models,~\cite{long2015fully} proposed fully convolutional networks for pixel-wise labeling. They proposed up-sampling (deconvolving) the output activation maps from which the pixel-wise output can be calculated. The overall architecture of the network is visualized in \review{Figure}~\ref{fully_conv}. 

\begin{figure}[!ht]
\centering
\includegraphics[width=.7\textwidth]{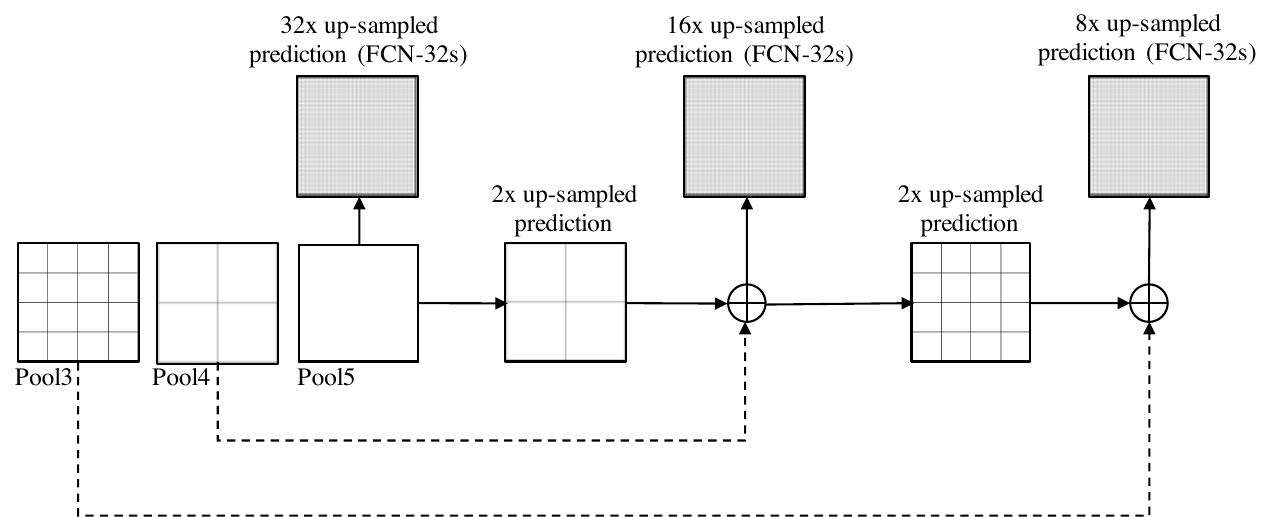}
\caption {Upsampling and fusion step of the fully convolution networks~\citep{long2015fully}.}
\label{fuse_fully_conv}
\end{figure}

In order to preserve the contextual  spatial information within an image as the filtered input data progresses deeper into the network, \cite{long2015fully} proposed to fuse the output with shallower layers' output. The fusion step is visualized in \review{Figure}~\ref{fuse_fully_conv}.

\subsection{Encoder-decoder Semantic Image Segmentation Networks}

\begin{figure}[!ht]
\centering
\includegraphics[width=.8\textwidth]{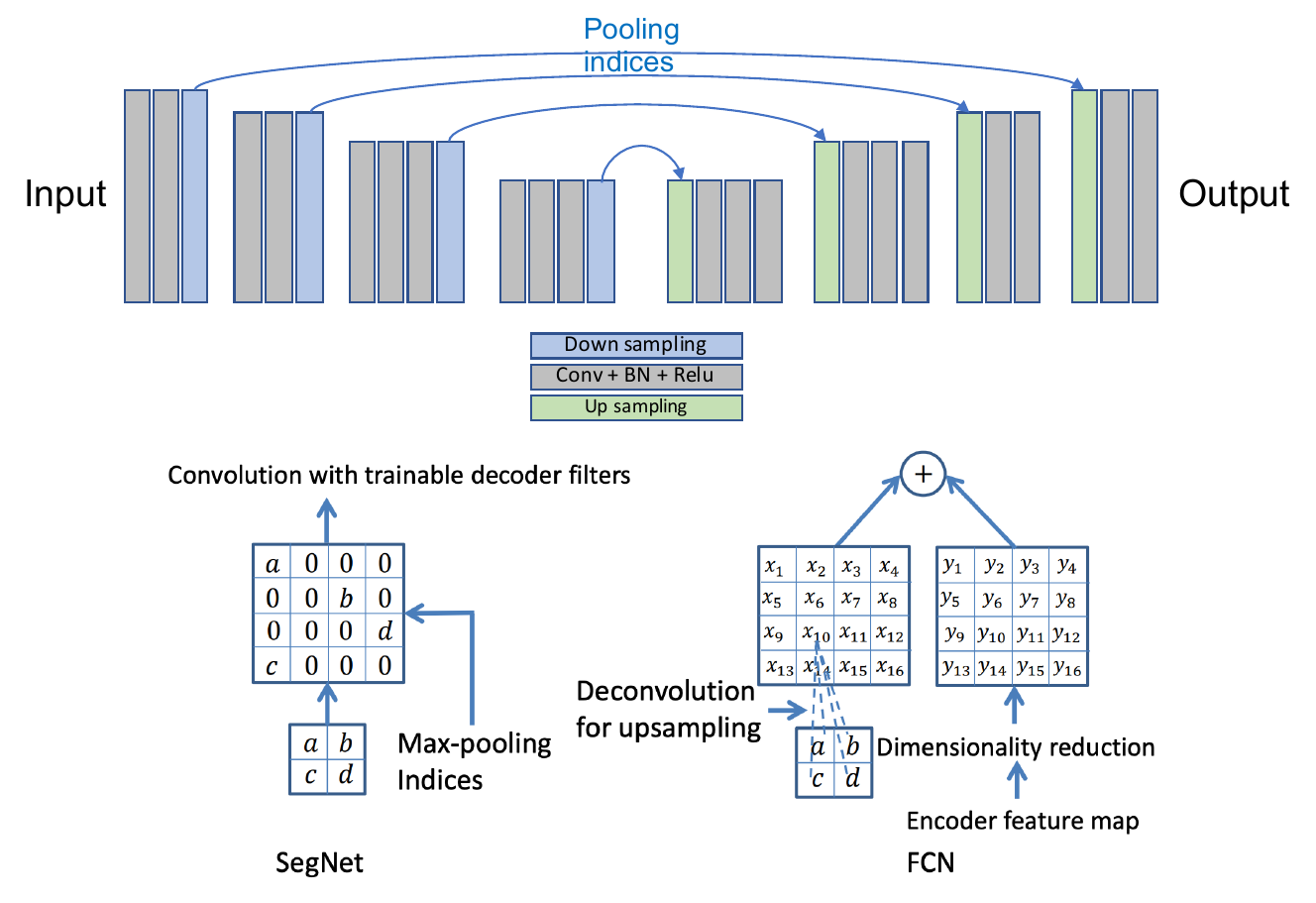}
\caption {\textbf{Top}: An illustration of the SegNet architecture. There are no fully connected layers, and hence it is only convolutional.  \textbf{Bottom}: An illustration of SegNet and FCN~\citep{long2015fully} decoders. $a, b, c, d$ correspond to values in a feature map. SegNet uses the max-pooling indices to upsample (without learning) the feature map(s) and convolves with a trainable decoder filter bank. FCN upsamples by learning to deconvolve the input feature map and adds the corresponding encoder feature map to produce the decoder output. This feature map is the output of the max-pooling layer (includes sub-sampling) in the corresponding encoder. Note that there are no trainable decoder filters in FCN~(\cite{badrinarayanan2015segnet}).}
\label{segnet}
\end{figure}

Next, encoder-decoder segmentation networks~\citep{noh2015learning} such as SegNet, were introduced~\citep{badrinarayanan2015segnet}. The role of the decoder network is to map the low-resolution encoder feature to full input resolution feature maps for pixel-wise classification. The novelty of SegNet lies in the manner in which the decoder upsamples the lower resolution input feature maps. Specifically, the decoder uses pooling indices (Figure~\ref{segnet}) computed in the max-pooling step of the corresponding encoder to perform non-linear upsampling. The architecture (\review{Figure}~\ref{segnet}) consists of a sequence of non-linear processing layers (encoder) and a corresponding set of decoder layers followed by a pixel-wise classifier. Typically, each encoder consists of one or more convolutional layers with batch normalization and a ReLU non-linearity, followed by non-overlapping max-pooling and sub-sampling. The sparse encoding due to the pooling process is upsampled in the decoder using the max-pooling indices in the encoding sequence.

\begin{figure}[!ht]
\centering
\includegraphics[width=.7\textwidth]{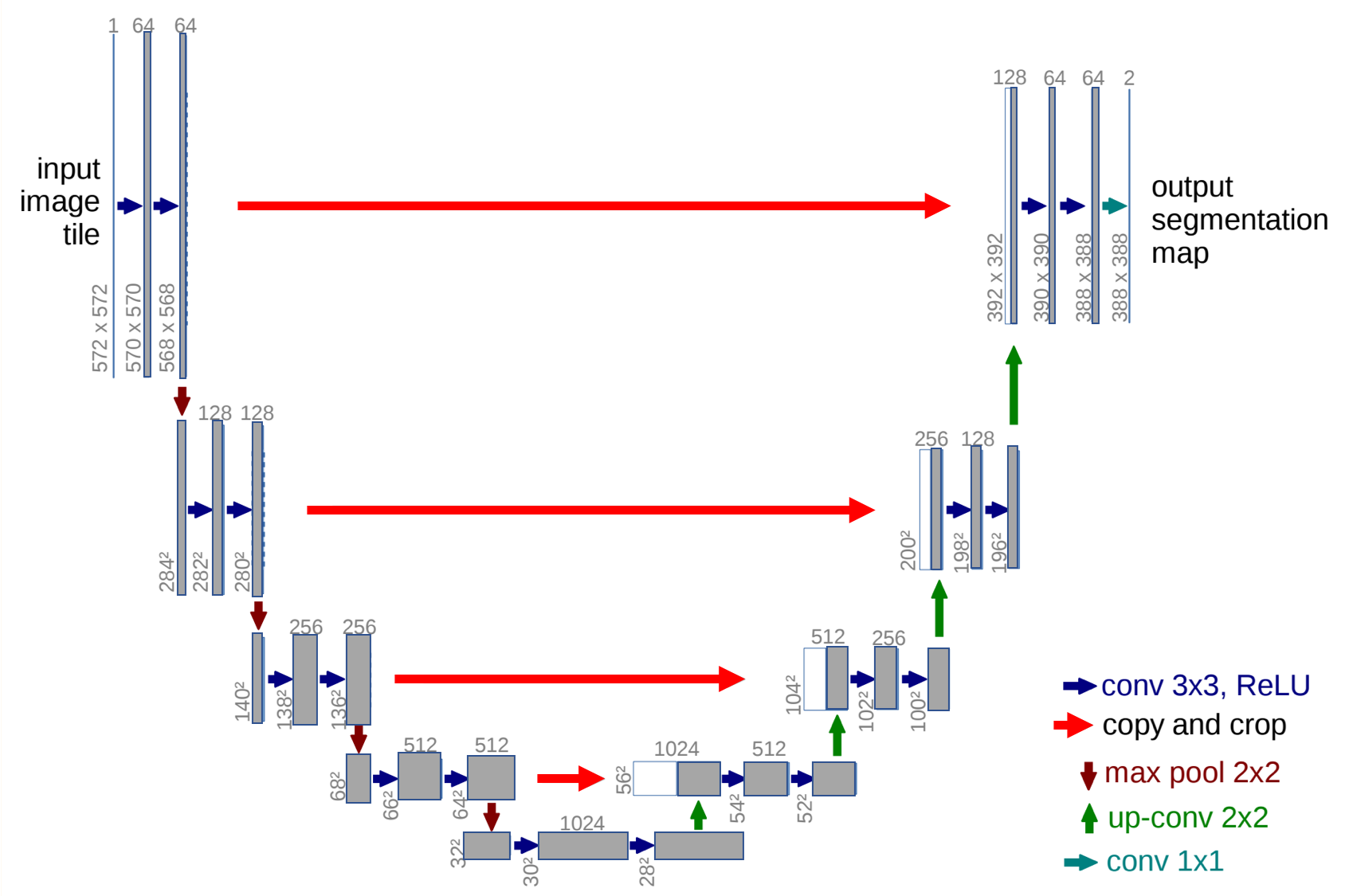}
\caption {An illustration of the U-Net~\citep{ronneberger2015u} architecture.}
\label{u-net}
\end{figure}

\cite{ronneberger2015u} proposed an architecture (U-Net; \review{Figure}~\ref{u-net}) consisting of a contracting path to capture context and a symmetric expanding path that enables precise localization. Similar to the image recognition~\citep{he2016deep} and keypoint detection~\citep{honari2016recombinator},~\cite{ronneberger2015u} added \textit{skip connections} to the encoder-decoder image segmentation networks, e.g., SegNet, which improved the model's accuracy and addressed the problem of vanishing gradients. 

\begin{figure}[!ht]
\centering
\includegraphics[width=.7\textwidth]{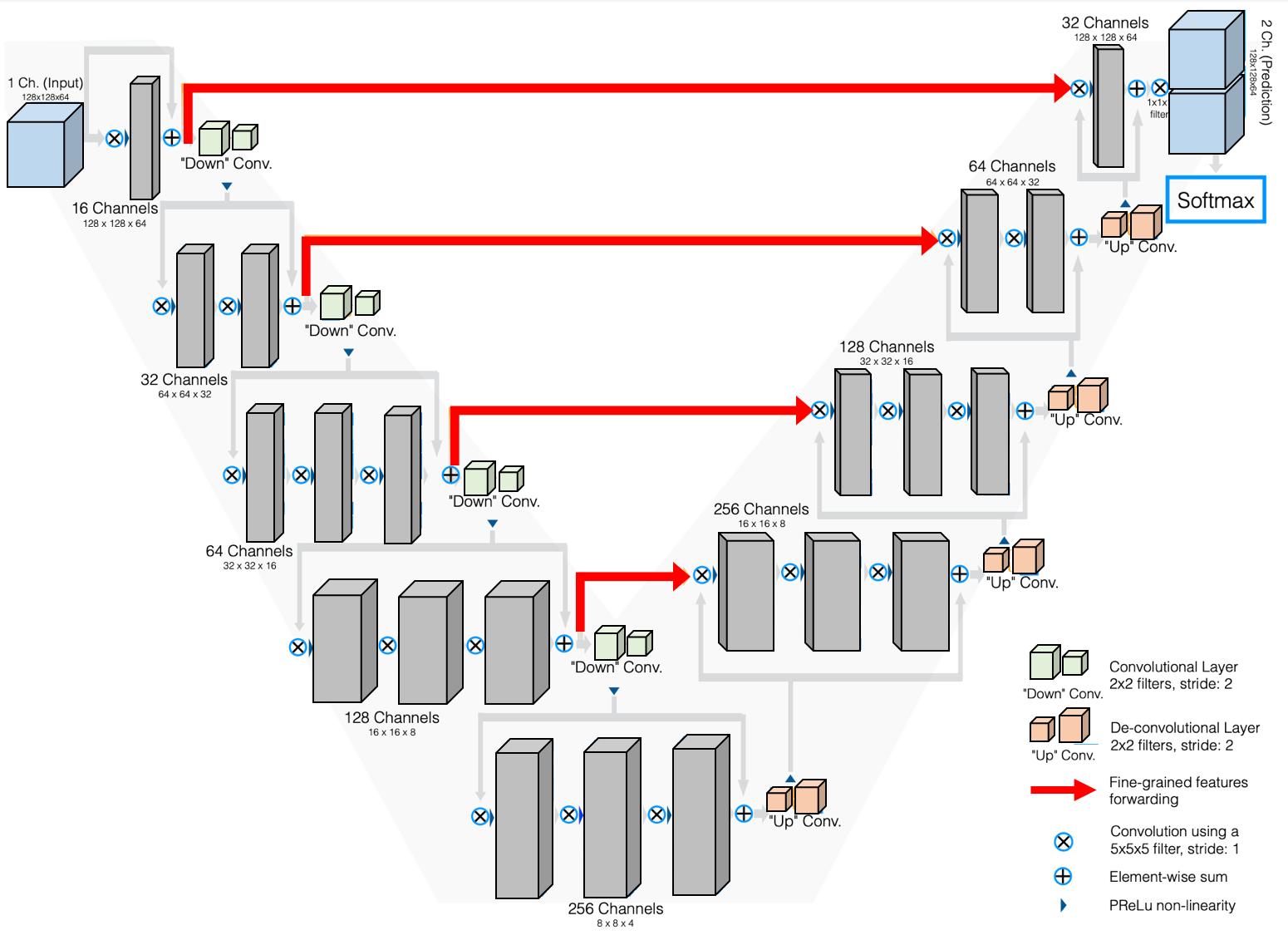}
\caption {An illustration of the V-Net~\citep{milletari2016v} architecture.}
\label{v-net}
\end{figure}

\cite{milletari2016v} proposed a similar architecture (V-Net; \review{Figure}~\ref{v-net}) which added residual connections and replaced 2D operations with their 3D counterparts in order to process volumetric images. Milletari et al. also proposed optimizing for a widely used segmentation metric, i.e., Dice, which will be discussed in more detail in the section~\ref{losses}.

\begin{figure}[!ht]
\centering
\includegraphics[width=.6\textwidth]{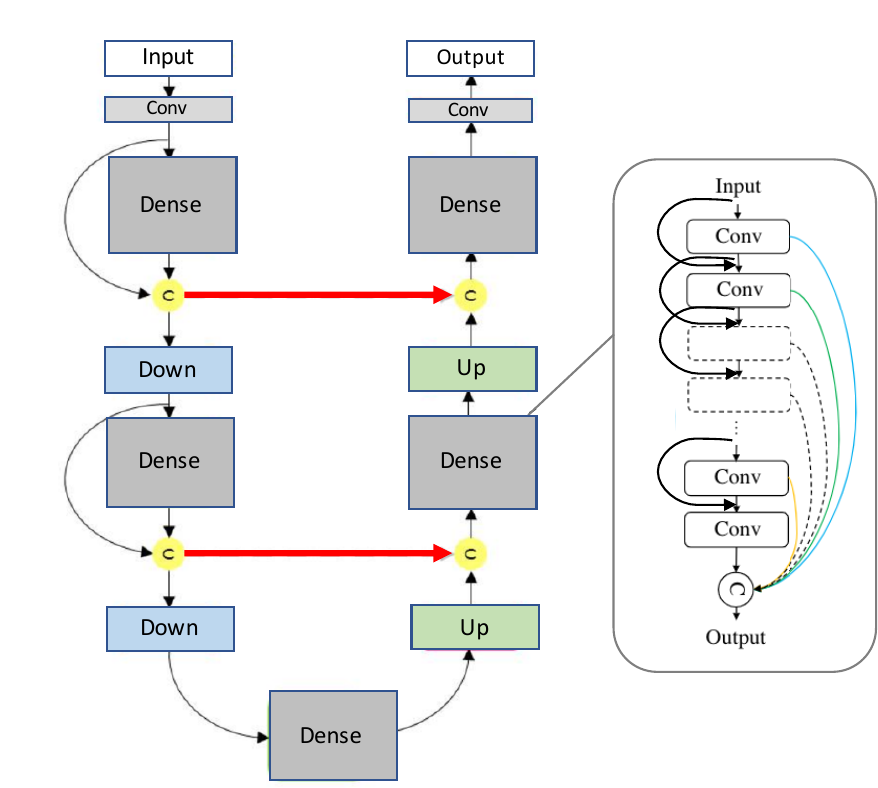}
\caption {Diagram of the one-hundred layers Tiramisu network architecture~\citep{jegou2017one}. The architecture is built from dense blocks. The architecture is composed of a downsampling path with two transitions down and an upsampling path with two transitions up. A circle represents concatenation, and the arrows represent connectivity patterns in the network. Gray horizontal arrows represent skip connections, where the feature maps from the downsampling path are concatenated with the corresponding feature maps in the upsampling path. Note that the connectivity pattern in the upsampling and the downsampling paths are different. In the downsampling path, the input to a dense block is concatenated with its output, leading to linear growth of the number of feature maps, whereas in the upsampling path, it is not the case.}
\label{tiramisu}
\end{figure}

\cite{jegou2017one} developed a segmentation version of the densely connected networks architecture (DenseNet~(\citet{huang2017densely}) by adapting the U-Net like encoder-decoder skeleton. In Figure~\ref{tiramisu}, the detailed architecture of the network is visualized. 

\begin{figure}[!ht]
\centering
\includegraphics[width=0.7\textwidth]{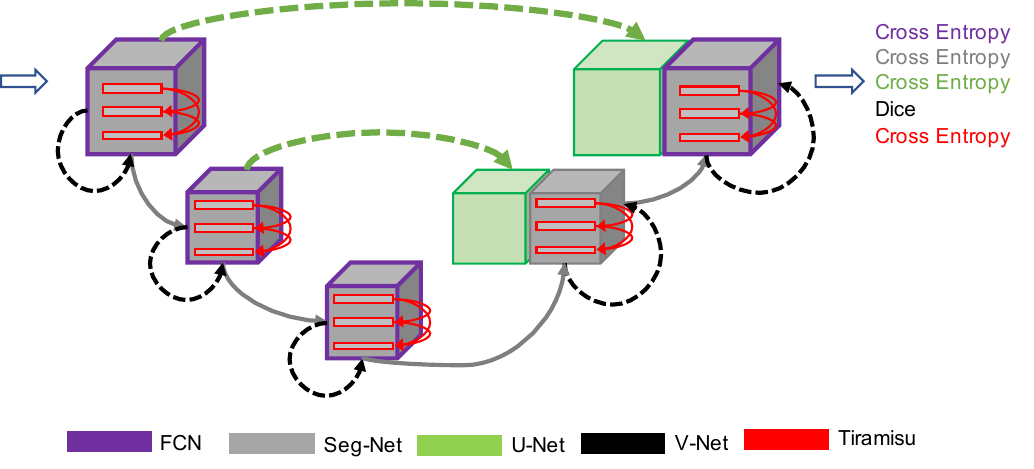}
\caption {Gradual architectural improvements applied to FCN~\citep{long2015fully} over time.}
\label{arc_all}
\end{figure}

In Figure~\ref{arc_all}, we visualize the \textit{simplified} architectural modifications applied to the first image segmentation network i.e. FCN.

\begin{figure}[!ht]
\centering
\includegraphics[width=.8\textwidth]{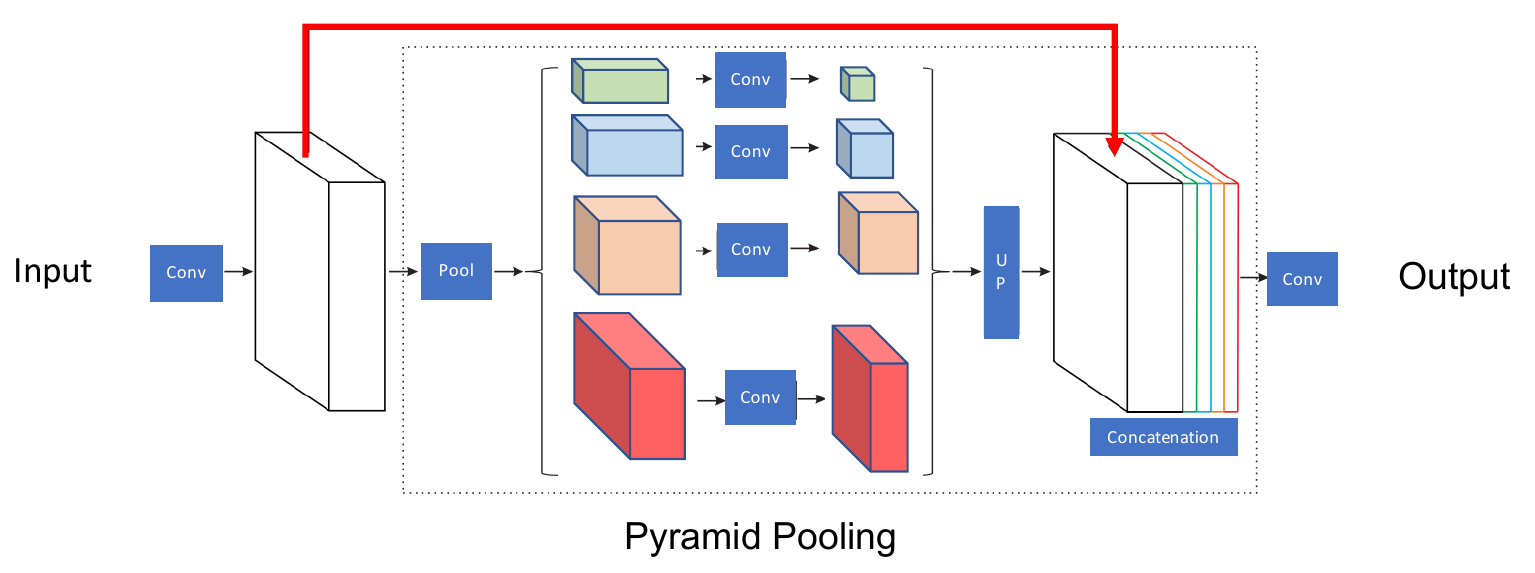}
\caption {Overview of the pyramid scene parsing networks. Given an input image (a), feature maps from last convolution layer are pulled (b), then a pyramid parsing module is applied to harvest different sub-region representations, followed by upsampling and concatenation layers to form the final feature representation, which carries both local and global context information in (c). Finally, the representation is fed into a convolution layer to get the final per-pixel prediction (d)~\cite{zhao2017pyramid}.}
\label{pspnet}
\end{figure}

Several modified versions (e.g. deeper/shallower, adding extra attention blocks) of encoder-decoder networks have been applied to semantic segmentation~\citep{amirul2017gated,fu2019stacked,lin2017refinenet,peng2017large,pohlen2017full,wojna2017devil,zhang2018exfuse}. Recently in 2018, DeepLabV3+~\citep{chen2018encoder} has outperformed many state-of-the-art segmentation networks on PASCAL VOC 2012~\citep{everingham2015pascal} and Cityscapes~\citep{cordts2016cityscapes} datasets.~\cite{zhao2017pyramid} modified the feature fusing operation proposed by~\cite{long2015fully} using a spatial pyramid pooling module or encode-decoder structure (Figure~\ref{pspnet}) are used in deep neural networks for semantic segmentation tasks. The spatial pyramid networks are able to encode multi-scale contextual information by probing the incoming features with filters or pooling operations at multiple rates and multiple effective fields-of-view, while the latter networks can capture sharper object boundaries by gradually recovering the spatial information. 

\begin{figure}[!ht]
\centering
\includegraphics[width=.8\textwidth]{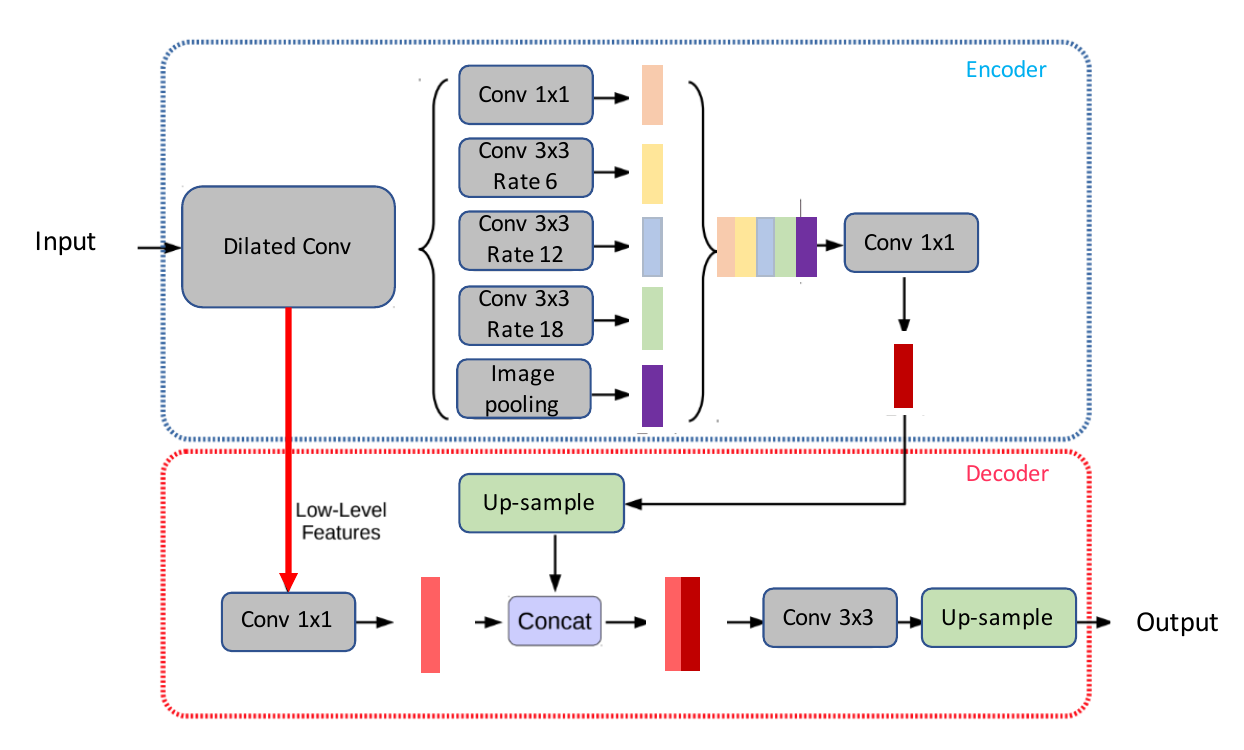}
\caption {An illustration of the DeepLabV3+; The encoder module encodes multi-scale contextual information by applying atrous (dilated) convolution at multiple scales, while the simple yet effective decoder module refines the segmentation results along object boundaries~\citep{chen2018encoder}.}
\label{deeplabv3+}
\end{figure}

\cite{chen2018encoder} proposed to combine the advantages from both dilated convolutions and feature pyramid pooling. Specifically, DeepLabv3+, extends DeepLabv3 (\cite{chen2017rethinking}) by adding a simple yet effective decoder module (\review{Figure}~\ref{deeplabv3+}) to refine the segmentation results, especially along object boundaries using dilated convolutions and pyramid features.

\subsection{Computation\review{al} Complexity Reduction for Image Segmentation Networks}

Several works have been done on reducing the time and the computational complexity of deep classification networks~\citep{howard2017mobilenets,leroux2018iamnn}. A few other works have attempted to simplify the structure of deep networks, e.g., by tensor factorization~\citep{kim2015compression}, channel/network pruning~\citep{wen2016learning}, or applying sparsity to connections~\citep{han2016eie}. \review{Similarly,~\cite{yu2018bisenet} addressed the high computational cost associated with high resolution feature maps in U-shaped architectures by proposing spatial and context paths to preserve the rich spatial information and obtain a large receptive field.} A few methods have focused on the complexity optimization of deep image segmentation networks. Similar to the work of~\cite{saxena2016convolutional},~\cite{liu2019auto} proposed a hierarchical neural architecture search for semantic image segmentation by performing both cell and network-level search and achieved comparable results to the state-of-the-art results on the PASCAL VOC 2012~\citep{everingham2015pascal} and Cityscapes~\citep{cordts2016cityscapes} datasets. In contrast,~\cite{chen2018searching} focused on searching the much smaller atrous spatial pyramid pooling module using random search. \review{Depth-wise separable convolutions~\citep{sifre2014rigid,chollet2017xception} offer computational complexity reductions since they have fewer parameters and have therefore also been used in deep segmentation models~\citep{chen2018encoder,sandler2018mobilenetv2}.}

Besides network architecture search,~\cite{srivastava2015highway} modified ResNet in a way to control the flow of information through a connection. \cite{lin2017refinenet} adopted one step fusion without filtering the channels. 

\subsection{Attention-based Semantic Image Segmentation}
Attention can be viewed as using information transferred from several subsequent layers/feature maps to select and localize the most discriminative (or salient) part of the input signal.~\cite{wang2017residual} added an attention module to the deep residual network (ResNet) for image classification. Their proposed attention module consists of several encoding-decoding layers. ~\cite{hu2017squeeze} proposed a selection mechanism where feature maps are first aggregated using global average pooling and reduced to a single channel descriptor. Then an activation gate is used to highlight the most discriminative features.~\review{\cite{wang2018non} proposed non-local operation blocks for encoding long range spatio-temporal dependencies with deep neural networks that can be plugged into existing architectures.}~\cite{fu2019dual} proposed dual attention networks that apply both spatial and channel-based attention operations.   

\cite{li2018pyramid} proposed a pyramid attention based network, for semantic segmentation. They combined an attention mechanism and a spatial pyramid to extract precise dense features for pixel labeling instead of complicated dilated convolution and artificially designed decoder networks.~\cite{chen2016attention} applied attention to DeepLab~\citep{chen2017deeplab} which takes multi-scale inputs.

\subsection{Adversarial Semantic Image Segmentation}
\cite{goodfellow2014generative} proposed an adversarial approach to learn deep generative models. Their generative adversarial networks (GANs) take samples $z$ from a fixed (e.g., standard Gaussian) distribution $p_{z}(z)$, and transform them using a deterministic differentiable deep network $p(.)$ to approximate the distribution of training samples $x$. Inspired by adversarial learning,~\cite{luc2016semantic} trained a convolutional semantic segmentation network along with an adversarial network that discriminates segmentation maps coming either from the ground truth or from the segmentation network. Their loss function is defined as

\begin{equation}
\begin{split}
    \ell\left(\boldsymbol{\theta}_{s}, \boldsymbol{\theta}_{a}\right) &=\sum_{n=1}^{N} \ell_{\mathrm{mce}}\left(s\left(\boldsymbol{x}_{n}\right), \boldsymbol{y}_{n}\right)\\
    & -\lambda\left[\ell_{\mathrm{bce}}\left(a\left(\boldsymbol{x}_{n}, \boldsymbol{y}_{n}\right), 1\right)\right. \left. + \ell_{\mathrm{bce}}\left(a\left(\boldsymbol{x}_{n}, s\left(\boldsymbol{x}_{n}\right)\right), 0\right)\right],
\end{split}
\end{equation}

\noindent where $\boldsymbol{\theta}_{s}$ and $\boldsymbol{\theta}_{a}$ denote the parameters of the segmentation and adversarial model, respectively. $l_{bce}$ and $l_{mce}$ are binary and multi-class cross-entropy losses, respectively. In this setup, the segmentor tries to produce segmentation maps that are close to the ground truth, i.e., which look more realistic. 

The main models being used for image segmentation mostly follow encoder-decoder architectures as U-Net. Recent approaches have shown that dilated convolutions and feature pyramid pooling can improve the U-Net style networks. In Section~\ref{arch-medical}, we summarize how these methods and their modified counterparts have been applied to medical images. 

\section{Architectural Improvements Applied to Medical Images} \label{arch-medical}
In this section, we review the different architectural based improvements for deep learning-based 2D and volumetric medical image segmentation.

\subsection{Model Compression based Image Segmentation}
To perform image segmentation in real-time and be able to process larger images or (sub) volumes in case of processing volumetric and high-resolution 2D images such as CT, MRI, and histopathology images, several methods have attempted to compress deep models.~\cite{weng2019unet} applied a neural architecture search method to U-Net to obtain a smaller network with a better organ/tumor segmentation performance on CT, MR, and ultrasound images.~\cite{brugger2019partially} by leveraging group normalization~\citep{wu2018group} and leaky ReLU function, redesigned the U-Net architecture in order to make the network more memory efficient for 3D medical image segmentation.~\cite{Perone2018-la} and~\cite{bonta2019efficient} designed a dilated convolution neural network with fewer parameters as compared to the original convolution-based one. Some other works~\citep{xu2018quantization,paschali2019deep} have focused on weight quantization of deep networks for making segmentation networks smaller. 

\subsection{Encoder Decoder based Image Segmentation}
\cite{drozdzal2018learning} proposed to normalize input images before segmentation by applying a simple CNN prior to pushing the images to the main segmentation network. They showed improved results on electron microscopy segmentation, liver segmentation from CT, and prostate segmentation from MRI scans.~\cite{gu2019net} proposed using a dilated convolution block close to the network's bottleneck to preserve contextual information. 

\cite{vorontsov2019boosting}, using a dataset defined in \cite{cohen2018distribution}, proposed an image-to-image based framework to transform an input image with object of interest (presence domain) like a tumor to an image without the tumor (absence domain) i.e. translate diseased image to healthy; next, their model learns to add the removed tumor to the new healthy image. This results in capturing detailed structure from the object, which improves the segmentation of the object.~\cite{zhou2018unet++} proposed a rewiring method for the long skip connections used in U-Net and tested their method on nodule segmentation in the low-dose CT scans of the chest, nuclei segmentation in the microscopy images, liver segmentation in abdominal CT scans, and polyp segmentation in colonoscopy videos.

\subsection{Attention based Image Segmentation}
\cite{nie2018asdnet} designed an attention model to segment prostate from MRI images with higher accuracy compared to baseline models, e.g.,
V-Net~\citep{milletari2016v} and FCN~\citep{long2015fully}.~\cite{sinha2019multi} proposed a multi-level attention based architecture for abdominal organ segmentation from MRI images. ~\cite{qin2018autofocus} proposed a dilated convolution base block to preserve more detailed attention in 3D medical image segmentation. Similarly, other papers~\citep{lian2018attention,isensee2018nnu,li2019attention,ni2019rasnet,oktay2018attention,schlemper2019attention} have leveraged the attention concept into medical image segmentation as well.

\subsection{Adversarial Training based Image Segmentation}
\cite{khosravan2019pan} proposed an adversarial training framework for pancreas segmentation from CT scans.~\cite{son2017retinal} applied \review{GAN}s for retinal image segmentation.~\cite{xue2018segan} used a  fully convolutional network as a segmenter in the generative adversarial framework to segment brain tumors from MRI images. Other papers~\citep{costa2017towards,dai2018scan,jin2018ct,moeskops2017adversarial,neff2017generative,rezaei2017conditional,yang2017automatic,zhang2017deep} have also successfully applied adversarial learning to medical image segmentation.

\subsection{Sequenced Models}
The Recurrent Neural Network (RNN) was designed for handling sequences. The long short-term memory (LSTM) network is a type of RNN that introduces self-loops to enable the gradient flow for long duration~\citep{hochreiter1997long}. In the medical image analysis domain, RNNs have been used to model the temporal dependency in image sequences.~\cite{bai2018recurrent} proposed an image sequence segmentation algorithm by combining a fully convolutional network with a recurrent neural network, which incorporates both spatial and temporal information into the segmentation task. Similarly,~\cite{gao2018fully} applied LSTM and CNN to model temporal relationship in brian MRI slices to improve segmentation performance in 4D volumes.~\cite{li2019pancreas} applied U-Net to obtain initial segmentation probability maps and further improve them using LSTM for pancreas segmentation from 3D CT scans. Similarly, other works have also applied RNNs (LSTMs)~\citep{alom2019recurrent,chakravarty2018race,yang2017fine,mengliu2019,mengliu2019mlmi} to medical image segmentation.

\section{Optimization Function based Improvements}\label{losses}
In addition to improved segmentation speed and accuracy using architectural modifications as mentioned in Section~\ref{netimprove}, designing new loss functions has also resulted in improvements in subsequent inference-time segmentation accuracy. 

\subsection{Cross Entropy}
The most commonly used loss function for the task of image segmentation is a pixel-wise cross entropy loss (Eqn. \ref{crossentropy}). This loss examines each pixel individually, comparing the class predictions vector to the one-hot encoded target (or ground  truth) vector. For the case of binary segmentation, let $P(Y=0)=p$ and $P(Y=1)=1-p$. The predictions are given by the logistic/sigmoid function  $P(\hat{Y}=0)=\frac{1}{1+e^{-x}}=\hat{p}$ and $P(\hat{Y}=1)=1-\frac{1}{1+e^{-x}}=1-\hat{p}$, where $x$ is output of network. Then cross entropy (CE) can be defined as:

\begin{equation} \label{crossentropy}
\operatorname{CE}(p, \hat{p})=-(p \log (\hat{p})+(1-p) \log (1-\hat{p})).\end{equation}

\noindent The general form of the equation for multi-region (or multi-class) segmentation can be written as:

\begin{equation} \label{mccrossentropy}
\operatorname{CE} = -\sum_{classes} p \log \hat{p}
\end{equation}

\subsection{Weighted Cross Entropy}
The cross-entropy loss evaluates the class predictions for each pixel vector individually and then averages over all pixels, which implies equal learning to each pixel in the image. This can be problematic if the various classes have unbalanced representation in the image, as the most prevalent class can dominate training.~\citet{long2015fully} discussed weighting the cross-entropy loss (WCE) for each class in order to counteract a class imbalance present in the dataset. WCE was defined as:

\begin{equation}
    \operatorname{WCE}(p, \hat{p})=-(\beta p \log (\hat{p})+(1-p) \log (1-\hat{p})).
\end{equation}

To decrease the number of false negatives, $\beta$ is set to a value larger than 1, and to decrease the number of false positives $\beta$ is set to a value smaller than 1. To weight the negative pixels as well, the following balanced cross-entropy (BCE) can be used~\citep{xie2015holistically}.

\begin{equation}
\operatorname{BCE}(p, \hat{p})=-(\beta p \log (\hat{p})+(1-\beta)(1-p) \log (1-\hat{p})).
\end{equation}

\cite{ronneberger2015u} added a distance function to the cross-entropy function to enforce learning distance between the components to enforce better segmentation in case of having very close objects to each other as follows:

\begin{equation}
\operatorname{BCE}(p, \hat{p})+w_{0} \cdot \exp \left(-\frac{\left(d_{1}(x)+d_{2}(x)\right)^{2}}{2 \sigma^{2}}\right)
\end{equation}

\noindent where $d_{1}(x)$ and $d_{2}(x)$ are two functions that calculate the distance to the border of nearest and second cells in their cell segmentation problem.

\subsection{Focal Loss}
To reduce the contribution of easy examples so that the CNN focuses more on the difficult examples,~\cite{lin2017focal} added the term $(1-\hat{p})^{\gamma}$ to the cross entropy loss as:

\begin{equation}    \label{focal_loss}
\begin{split}
    \operatorname{FL}(p, \hat{p})=-\left(\alpha(1-\hat{p})^{\gamma} p \log (\hat{p})\right.\left.+(1-\alpha) \hat{p}^{\gamma}(1-p) \log (1-\hat{p})\right).
\end{split}
\end{equation}

\noindent Setting  $\gamma = 0$ in this equation yields the BCE loss. 

\subsection{Overlap Measure based Loss Functions}
\subsubsection{Dice Loss / F1 Score}
Another popular loss function for image segmentation tasks is based on the Dice coefficient, which is essentially a measure of overlap between two samples and is equivalent to the F1 score. This measure ranges from 0 to 1, where a Dice coefficient of 1 denotes perfect and complete overlap. The Dice coefficient (DC) is calculated as:

\begin{equation} \label{dice-coef}
\mathrm{DC}=\frac{2 T P}{2 T P+F P+F N}=\frac{2|X \cap Y|}{|X|+|Y|}.
\end{equation}

\noindent Similarly, the Jaccard metric (intersection over union: IoU) is computed as:

\begin{equation} \label{jac}
\mathrm{IoU}=\frac{T P}{T P+F P+F N}=\frac{|X \cap Y|}{|X|+|Y|-|X \cap Y|}
\end{equation}

\noindent where $X$ and $Y$ are the predicted and ground truth segmentation, respectively. TP is the true positives, FP false positives and FN false negatives. We can see that $\mathrm{DC} \geq \mathrm{IoU}$.

To use this as a loss function the DC can be defined as a Dice loss (DL)  function~\citep{milletari2016v}:

\begin{equation} \label{dice}
\mathrm{DL}(p, \hat{p})=\frac{2\langle p, \hat{p}\rangle}{\|p\|_{1}+\|\hat{p}\|_{1}}
\end{equation}

\noindent where $p \in\{0,1\}^{n} \text { and } 0 \leq \hat{p} \leq 1$. $p$ and $\hat{p}$ are the ground truth and predicted segmentation and ${\langle\cdot,\cdot\rangle}$ denotes dot product.

\subsubsection{Tversky Loss}
Tversky loss (TL)~\citep{salehi2017tversky} is a generalization of the DL. To control the level of FP and FN, TL weights them as the following:

\begin{equation} \label{Tversky}
\mathrm{TL}(p, \hat{p})=\frac{\langle p, \hat{p} \rangle}{\langle p, \hat{p}\rangle + \beta(1-p, \hat{p}\rangle+(1-\beta)(p, 1-\hat{p})}
\end{equation}

\noindent setting $\beta = 0.5$ simplifies the equation to Eqn. \ref{dice}.

\subsubsection{Exponential Logarithmic Loss}

\cite{wong20183d} proposed using a weighted sum of the exponential logarithmic Dice loss ($\mathcal{L}_{\mathrm{eld}}$) and the weighted exponential cross-entropy loss ($\mathcal{L}_{\mathrm{wece}}$) in order to improve the segmentation accuracy on small structures for tasks where there is a large variability among the sizes of the objects to be segmented. 
\begin{equation}    \label{el_loss}
    \mathcal{L} = w_{\mathrm{eld}}\mathcal{L}_{\mathrm{eld}} + w_{\mathrm{wece}}\mathcal{L}_{\mathrm{wece}},
\end{equation}
where
\begin{equation}
    \mathcal{L}_{\mathrm{eld}} = \text{\textbf{E}}\left[ \left( -\ln{(D_i)} \right)^{\gamma_D} \right], \ \text{and}
\end{equation}
\begin{equation}
    \mathcal{L}_{\mathrm{wece}} = \text{\textbf{E}}\left[ \left( -\ln{(p_l(\textbf{x}))} \right)^{\gamma_{CE}} \right].
\end{equation}
\textbf{x}, $i$, and $l$ denote the pixel position, the predicted label, and the ground truth label. $D_i$ denotes the smoothed Dice loss (obtained by adding an $\epsilon = 1$ term to the numerator and denominator in Eqn.~\ref{dice} in order to handle missing labels while training, and $\gamma_D$ and $\gamma_{CE}$ are used to control the non-linearities of the respective loss functions.

\subsubsection{Lov{\'a}sz-Softmax loss}

Since it has been shown that the Jaccard loss (IoU loss) is submodular~\citep{berman2018yes},~\cite{matthew2018lovasz} proposed using the Lov{\'a}sz hinge with the Jaccard loss for binary segmentation, and proposed a surrogate of the Jaccard loss, called the Lov{\'a}sz-Softmax loss, which can be applied for the multi-class segmentation task. The Lov{\'a}sz-Softmax loss is, therefore, a smooth extension of the discrete Jaccard loss, and is defined as
\begin{equation}
    \mathcal{L}_{\mathrm{LovaszSoftmax}} = \dfrac{1}{|\mathcal{C}|} \sum_{c\in\mathcal{C}}\overline{\Delta_{J_c}}\left(\bm{m}(c)\right),
\end{equation}

where ${\Delta_{J_c}}\left(\cdot\right)$ denotes the convex closure of the submodular Jaccard loss, $\overline{\cdot}$ denotes that it is a tight convex closure and polynomial time computable, $\mathcal{C}$ denotes all the classes, and ${J_c}$ and $\bm{m}(c)$ denote the Jaccard index and the vector of errors for class $c$ respectively.

\subsubsection{Boundary Loss}

\cite{pmlr-v102-kervadec19a} proposed to calculate boundary loss $\mathcal{L}_{B}$ along with the generalized Dice loss $\mathcal{L}_{GD}$ function as 

\begin{equation} \label{boundry1}
\alpha \mathcal{L}_{GD}(\theta)+(1-\alpha) \mathcal{L}_{B}(\theta),
\end{equation}
\noindent where the two terms in the loss function are defined as

\begin{equation}
\begin{split}
    \mathcal{L}_{G D}(\theta) &= 1 \ - 2\dfrac{ w_{G} \sum_{p \in \Omega} g(p) s_{\theta}(p) + w_{B} \sum_{p \in \Omega}(1-g(p))\left(1-s_{\theta}(p)\right) }
    { w_{G} \sum_{p \in \Omega}\left[s_{\theta}(p)+g(p)\right] + w_{B} \sum_{p \in \Omega}\left[2-s_{\theta}(p)-g(p)\right] }, \ \text{and}
\end{split}
\end{equation}

\normalsize
\begin{equation} \label{boundry3}
\mathcal{L}_{B}(\theta) = {p \in \Omega} \phi_{G}(p) s_{\theta}(p),
\end{equation}

\noindent where $\phi_{G}(p)=-\left\|p-z_{\partial G}(p)\right\|$ if $p \in G$ and 
$\phi_{G}(p)=\left\|p-z_{\partial G}(p)\right\|$, otherwise. The general form integral 
$\sum_{\Omega} g(p) f\left(s_{\theta}(p)\right)$ is for foreground and 
$\sum_{\Omega}(1-g(p)) f\left(1-s_{\theta}(p)\right)$ for background. 
$w_{G}=1 /\left(\sum_{p \in \Omega} g(p)\right)^{2}$ and \\
$w_{B}=1 /\left(\sum_{\Omega}(1-g(p))\right)^{2}.\Omega$ shows the spatial domain.

\subsubsection{Conservative Loss}

\cite{zhu2018penalizing} proposed the Conservative Loss for in order to achieve a good generalization ability in domain adaptation tasks by penalizing the extreme cases and encouraging the moderate cases. The Conservative Loss is defined as
\begin{equation}
    CL(p_t) = \lambda(1 + \log_a(p_t))^2 * \log_a(-\log_a(p_t)),
\end{equation}
where $p_t$ is the probability of the prediction towards the ground truth and $a$ is the base of the logarithm. $a$ and $\lambda$ are empirically chosen to be $e$ (Euler's number) and 5 respectively.



Other works also include approaches to optimize the segmentation metrics~\citep{nowozin2014optimal}, weighting the loss function~\citep{roy2017error}, and adding regularizers to loss functions to encode geometrical and topological shape priors~\citep{aicha_shapepriors,mirikharaji2018star}.

A significant problem in image segmentation (particularly in medical images) is to overcome class imbalance for which overlap measure based methods have shown reasonably good performance in overcoming the imbalance. In Section~\ref{loss_med}, we summarize the approaches which use new loss functions, particularly for medical image segmentation or use the (modified) loss functions mentioned above.

\begin{figure*}[!t]
\centering
\includegraphics[width=\textwidth]{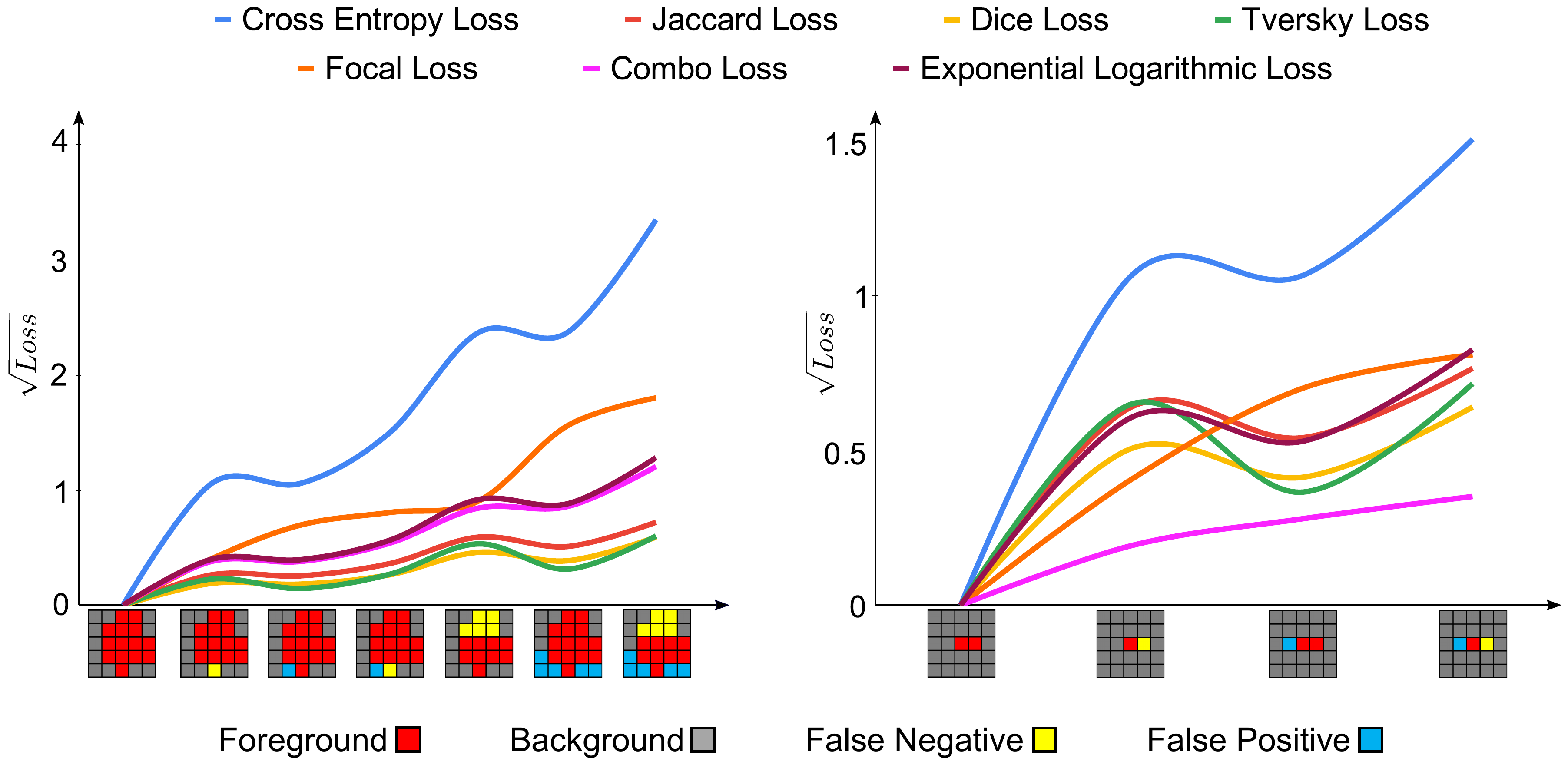}
\caption {A comparison of seven loss functions for different extends of overlaps for a large (left) and a small (right) object.}
\label{loss_trends}
\end{figure*}

\begin{figure}[h!]
    \centering
    \includegraphics[scale=.3]{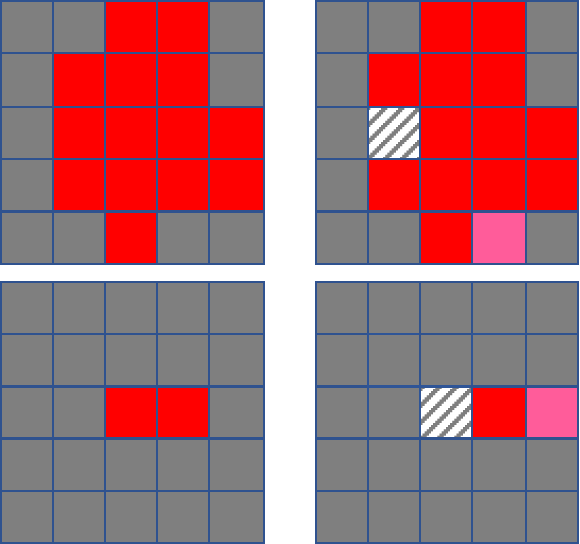}
    \caption {Comparison of cross entropy and Dice losses for segmenting small and large objects. The red pixels show the ground truth and the predicted foregrounds in the left and right columns respectively. The striped and the pink pixels indicate false negative and false positive, respectively. For the top row (i.e., large foreground), the Dice loss returns $0.96$ for one false negative and for the bottom row (i.e., small object) returns $0.66$ for one false negative, whereas the cross entropy loss function outputs $0.83$ for both the cases. By considering a false negative and false positive, the output value drops even more in case of using Dice but the cross entropy stays smooth (i.e., Dice value of $0.93$ and $0.50$ for large and small object versus cross entropy loss value of $1.66$ for both.)}
    \label{ce_dice}
\end{figure}

In Figure~\ref{loss_trends}, we visualize the behavior of different loss functions for segmenting large and small objects. For the parameters of the loss functions, we use the same parameters as reported by the authors in their respective papers. Therefore, we use $\beta=0.3$ in Eqn.~\ref{Tversky}, $\alpha=0.25$ and $\gamma=2$ in Eqn.~\ref{focal_loss}, and $\gamma_D = \gamma_{CE} = 1$, $w_{\mathrm{eld}}=0.8$, and $w_{\mathrm{wece}}=0.2$ in Eqn.~\ref{el_loss}. Moving from the left to the right for each plot, the overlap of the predictions and ground truth mask becomes progressively smaller, i.e., producing more false positives and false negatives. Ideally, the loss value should monotonically increase as more false positives, and negatives are predicted. For large objects, almost all the functions follow this assumption; however, for the small objects (right plot), only combo loss and focal loss penalize \textit{monotonically} more for larger errors. In other words, the overlap-based functions highly fluctuate while segmenting small and large objects (also see \review{Figure}~\ref{ce_dice}), which results in unstable optimization. The loss functions which use cross-entropy as the base and the overlap measure functions as a weighted regularizer show more \textit{stability} during training.

\section{Optimization Function based Improvements Applied to Medical Images} \label{loss_med}
The standard CE loss function and its weighted versions, as discussed in Section~\ref{losses}, have been applied to numerous medical image segmentation problems~\citep{isensee2018nnu,li2019attention,lian2018attention,ni2019rasnet,nie2018asdnet,oktay2018attention,schlemper2019attention}. However,~\cite{milletari2016v} found that optimizing \review{CNNs} for DL (Eqn.~\ref{dice}) in some cases, e.g., in the case of having very small foreground objects in a large background, works better than the original cross-entropy.

\cite{li2019transformation} proposed adding the following regularization term to the cross entropy loss function to encourage smooth segmentation outputs.

\begin{equation} \label{smoothterm}
R=\sum_{i=1}^{N} \mathbb{E}_{\xi^{\prime}, \xi}\left\|f\left(x_{i} ; \theta, \xi^{\prime}\right)-f\left(x_{i} ; \theta, \xi\right)\right\|^{2}
\end{equation}

\noindent where $\xi^{\prime}$ and $\xi$ are different perturbation (e.g., Gaussian noise, network dropout, and randomized data
transformation) applied to the input image $x_i$.

\cite{chen2019learning} proposed leveraging traditional active contour energy minimization into \review{CNNs} via the following loss function. 

\begin{equation} \label{ac0}
\operatorname{Loss} = \operatorname{Length}+\lambda \cdot \operatorname{Region}
\end{equation}

\begin{equation} \label{ac1}
\text {Length}=\sum_{\Omega}^{i=1, j=1} \sqrt{\left|\left(\nabla u_{x_{i, j}}\right)^{2}+\left(\nabla u_{y_{i, j}}\right)^{2}\right|+\epsilon}
\end{equation}

\noindent where $x$ and $y$ from $u_{x_{i, j}}$ and $u_{y_{i, j}}$ are horizontal and vertical directions, respectively.

\begin{equation} \label{ac2}
\begin{split} \text {Region}=\left|\sum_{\Omega}^{i=1, j=1} u_{i, j}\left(c_{1}-v_{i, j}\right)^{2}\right| +\left|\sum_{\Omega}^{i=1, j=1}\left(1-u_{i, j}\right)\left(c_{2}-v_{i, j}\right)^{2}\right| \end{split}
\end{equation}

\noindent where $u$ and $v$ are represented as
prediction and a given image, respectively. c1 is set to 1 and c2 to 0. 
Similar to,~\cite{li2019transformation},~\cite{zhou2019high} proposed adding a contour regression term to the weighted cross entropy loss function. 

\cite{karimi2019reducing} optimized Hausdorff distance based function between a predicted and ground truth segmentation as follows.

\begin{equation} \label{hd1}
f_{\mathrm{HD}}(p, q)=\operatorname{Loss}(p, q)+\lambda\left(1-\frac{2 \sum_{\Omega}(p \circ q)}{\sum_{\Omega}\left(p^{2}+q^{2}\right)}\right)
\end{equation}

\noindent where the second term is the Dice loss function and the first term can be replaced with three different versions of the Hausdorff distance for $p$ and $q$ i.e. ground truth and predicted segmentations respectively, as follows;

\begin{equation} \label{hd2}
\operatorname{Loss}(q, p)=\frac{1}{|\Omega|} \sum_{\Omega}\left((p-q)^{2} \circ\left(d_{p}^{\alpha}+d_{q}^{\alpha}\right)\right)
\end{equation}

The parameter $\alpha$ determines the level of penalty for larger errors. $d_p$ is the distance map of the ground-truth segmentation as the unsigned distance to the boundary $\delta p$. Similarly, $d_q$ is defined as the distance to $\delta q$. The $\circ$ is Hadamard operation.

\begin{equation} \label{hd3}
\operatorname{Loss}(q, p)=\frac{1}{|\Omega|} \sum_{k=1}^{K} \sum_{\Omega}\left((p-q)^{2} \ominus_{k} B\right) k^{\alpha}
\end{equation}

\noindent where $\ominus_{k}$ denotes $k$ successive erosions. where
\begin{equation}
    B=\left(\begin{array}{ccc}{0} & {1 / 5} & {0} \\ {1 / 5} & {1 / 5} & {1 / 5} \\ {0} & {1 / 5} & {0}\end{array}\right)
\end{equation}

\begin{equation} \label{hd4}
\begin{split}
    \operatorname{Loss}(q, p)&=\frac{1}{|\Omega|} \sum_{r \in R} r^{\alpha} \sum_{\Omega}\left[f_{s}\left(B_{r} * \overline{p}^{C}\right) \circ f_{\overline{q} \backslash \overline{p}}\right.+ {f_{s}\left(B_{r} * \overline{p}\right) \circ f_{\overline{p} \backslash \overline{q}}} \\&+ {f_{s}\left(B_{r} * \overline{q}^{C}\right) \circ f_{\overline{p} \backslash \overline{q}}} + {f_{s}\left(B_{r} * \overline{q}\right) \circ f_{\overline{q} \backslash \overline{p}} ]}
\end{split}
\end{equation}

\noindent where $f_{\overline{q} \backslash \overline{p}}=(p-q)^{2} q$. $f_s$ indicates soft thresholding. $B_r$ denotes a circular-shaped convolutional kernel of radius r. Elements of $B_r$ are normalized such that they sum to one. $\overline{p}^{C}=1-\overline{p}$. Ground-truth and predicted segmentations, denoted with $\overline{p}$ and $\overline{q}$,

\cite{caliva2019distance} proposed to measure distance of each voxel to the boundaries of the objects and use the weight matrices to penalize a model for error on the boundaries.~\cite{kim2019multiphase} proposed using level-set energy minimization as a regularizer summed with standard multi-class cross entropy loss function for semi-supervised brain MRI segmentation as:

\begin{equation}
    \begin{aligned} \mathcal{L}_{\text {level}}(\Theta ; x)= \sum_{n=1}^{N} \int_{\Omega}\left|x(r)-c_{n}^{\Theta}\right|^{2} y_{n}^{\Theta}(r) d r +\lambda \sum_{n=1}^{N} \int_{\Omega}\left|\nabla y_{n}^{\Theta}(r)\right| d r \end{aligned}
\end{equation}

with 

\begin{equation}
    c_{n}^{\Theta}=\frac{\int_{\Omega} x(r) y_{n}^{\Theta}(r) d r}{\int_{\Omega} y_{n}^{\Theta}(r) d r}
\end{equation}

where  $x(r)$  is the input,  $y_{n}^{\Theta}(r)$ is the output of softmax layer, $\Theta$ refers to learnable parameters. 

\cite{taghanaki2019combo} discussed the risks of using solo overlap based loss functions and proposed to use them as regularizes along with a weighted cross entropy to explicitly handle input and output imbalance as follows;

\begin{multline}
    Combo \ Loss= \alpha \biggl(-\frac{1}{N} \sum_{i=1}^{N} \beta \left(t_i - \ln  p_i \right) + \left(1-\beta \right) \left[\left(1-t_i\right) \ln \left(1-p_i \right)\right]\biggr) \\+ \left ( 1-\alpha \right) \sum_{i=1}^{K} \left ( -\frac{2\sum_{i=1}^{N} p_i  t_i + S}{\sum_{i=1}^{N} p_i  + \sum_{i=1}^{N} t_i + S} \right)
\end{multline}

\noindent where $\alpha$ controls the amount of Dice term contribution in the loss function $L$, and $\beta \in [0,1]$ controls the level of model penalization for false positives/negatives: when $\beta$ is set to a value smaller than 0.5, $FP$ are penalized more than $FN$ as the term $(1-t_{i}) \ \ln \ (1-p_{i})$ is weighted more heavily, and vice versa. In their implementation, to prevent division by zero, the authors perform add-one smoothing (a specific instance of the additive/Laplace/Lidstone smoothing;~\citet{russell2016artificial}), i.e., they add unity constant $S$ to both the denominator and numerator of the Dice term. 

The majority of the methods discussed in Section~\ref{loss_med} have attempted to handle the class imbalance issue in the input images i.e., small foreground versus large background with providing weights/penalty terms in the loss function. Other approaches consist of first identifying the object of interest, cropping around this object, and then performing the task (e.g., segmentation) with better-balanced classes. This type of cascade approach has been applied for the segmentation of multiple sclerosis lesions in the spinal cord~\citep{Gros2018-ss}.

\section{Image Synthesis based Methods}\label{synthesis}


\review{Deep CNNs are heavily reliant on big data to avoid overfitting and class imbalance issues, and therefore this section focuses on data augmentation, a data-space solution to the problem of limited data. Apart from standard online image augmentation methods such as geometric transformations~\citep{lecun1998gradient,simard2003best,cirecsan2011high,cirecsan2012multi,24}, color space augmentations~\citep{galdran2017data,yading2017automatic,abhishek2020illumination}, etc., in this section, we discuss image synthesis methods, the output of which are novel images rather than modifications to existing images. GANs based augmentation techniques for segmentation tasks have been used for a wide variety of problems - from remote sensing imagery~\citep{mohajerani2019cloudmaskgan} to filamentary anatomical structures~\citep{zhao2017synthesizing}. For a more detailed review of image augmentation strategies in deep learning, we direct the interested readers to~\cite{Shorten2019}.}

\subsection{\review{Image Synthesis based Methods Applied to Natural Image Segmentation}}\label{natural_synthesis}

\review{\cite{neff2018generative} trained a Wasserstein GAN with gradient penalty~\citep{gulrajani2017improved} to generate labeled image data in the form of image-segmenation mask pairs. They evaluated their approach on a dataset of chest X-ray images and the Cityscapes dataset, and found that the WGAN-GP was able to generate images with sufficient variety and that a segmentation model trained using GAN-based augmentation only was able to perform better than a model trained with geometric transformation based augmentation. \cite{cherian2019semantically} proposed to incorporate semantic consistency in image-to-image translation task by introducing segmentation functions in the GAN architecture and showed that the semantic segmentation models trained with synthetic images led to considerable performance improvements. Other works include GAN-based data augmentation for domain adaptation~\citep{huang2018auggan,choi2019self} and panoptic data augmentation~\citep{liu2019panda}. However, the majority of GAN based data augmentation has been applied to medical images~\citep{Shorten2019}. Next, we discuss the GAN based image synthesis for augmentation in the field of medical image analysis.}

\subsection{Image Synthesis based Methods Applied to Medical Image Segmentation}\label{medical_synthesis}

\cite{chartsias2017adversarial} used a conditional GAN to generate cardiac MR images from CT images. They showed that utilizing the synthetic data increased the segmentation accuracy and that using only the synthetic data led to only a marginal decrease in the segmentation accuracy. Similarly,~\cite{zhang2018translating} proposed a GAN based volume-to-volume translation for generating MR volumes from corresponding CT volumes and vice versa. They showed that synthetic data improve segmentation performance on cardiovascular MRI volumes.~\cite{huo2018adversarial} proposed an end-to-end synthesis and segmentation network called EssNet to simultaneously synthesize CT images from unpaired MR images and to segment CT splenomegaly on unlabeled CT images and showed that their approach yielded better segmentation performance than even segmentation obtained using models trained using the manual CT labels.~\cite{abhishek2019mask2lesion} trained a conditional GAN to generate skin lesion images from and confined to binary masks, and showed that using the synthesized images led to a higher skin lesion segmentation accuracy.~\cite{zhang2018task} trained a GAN for translating between digitally reconstructed radiographs and X-ray images and achieved similar accuracy as supervised training in multi-organ segmentation.~\cite{shin2018medical} proposed a method to generate synthetic abnormal MRI images with brain tumors by training a \review{GAN} using two publicly available data sets of brain MRI. Similarly, other works~\citep{han2019learning,yang2018mri,yu20183d} have leveraged GANs to synthesize brain MR images.

\section{Weakly Supervised Methods}
Collecting large-scale accurate pixel-level annotation is time-consuming and financially expensive. However, unlabeled and weakly-labeled images can be collected in large amounts in a relatively fast and cheap manner. \review{As shown in Figure~\ref{fig:overview}, varying levels of supervision are possible when training deep segmentation models, from pixel-wise annotations (supervised learning) and image-level and bounding box annotations (semi-supervised learning) to no annotations at all (unsupervised learning), the last two of which comprise weak supervision.} Therefore, a promising direction for semantic image segmentation is to develop weakly supervised segmentation models.  

\subsection{\review{Weakly Supervised Methods Applied to Natural Images}}

\review{\cite{Kim2016ScaleInvariantFL} proposed a weakly supervised semantic segmentation network using unpooling and deconvolution operations, and used feature maps from the deconvolutions layers to learn scale-invariant features, and evaluated their model on the PASCAL VOC and chest X-ray image datasets.~\cite{lee2019ficklenet} used dropout~\citep{srivastava2014dropout} to choose features at random during training and inference and combine the many different localization maps to generate a single localization map, effectively discovering relationships between locations in an image, and evaluated their proposed approach on the PASCAL VOC dataset.}

\subsection{Weakly Supervised Methods Applied to Medical Images}

The scarcity of richly annotated medical images is limiting supervised deep learning-based solutions to medical image analysis tasks~\citep{Perone2019-jt}, such as localizing discriminatory radiomic disease signatures. Therefore, it is desirable to leverage unsupervised and weakly supervised models.
~\cite{KERVADEC201988} introduced a differentiable term in the loss function for datasets with weakly supervised labels, which reduced the computational demand for training while also achieving almost similar performance to full supervision for segmentation of cardiac images.~\cite{afshari2019weakly} used a fully convolutional architecture along with a Mumford-Shah functional~\cite{mumford1989optimal} inspired loss function to segment lesions from PET scans using only bounding box annotations as supervision.~\cite{mirikharaji2019learning} proposed to learn spatially adaptive weight maps to account for spatial variations in pixel-level annotations and used noisy annotations to train a segmentation model for skin lesions.~\cite{taghanaki2019infomask} proposed to learn spatial masks using only image-level labels with minimizing mutual information between the input and masks, and at the same time maximizing the mutual information between the masks and image labels.~\cite{peng2019discretely} proposed an approach to train a CNN with discrete constraints and regularization priors based on the alternating direction method of multipliers (ADMM).~\cite{Perone2018-lo} expanded the semi-supervised mean teacher~\citep{Tarvainen2017-zn} approach to segmentation tasks on MRI data, and show that it can bring important improvements in a realistic small data regime. In another work,~\cite{Perone2019-eo} extended the method of unsupervised domain adaptation using self-ensembling for the semantic segmentation task. They showed how this approach could improve the generalization of the models even when using a small amount of unlabeled data.

\section{Multi-Task Models}
Multi-task learning~\citep{caruana1997multitask} refers to a machine learning approach where multiple tasks are learned simultaneously, and the learning efficiency and the model performance on each of the tasks are improved because of the existing commonalities across the tasks. \review{For visual recognition tasks, it has been shown that there exist relations between various tasks in the task space~\citep{zamir2018taskonomy}, and multi-task models can help exploit these relationships to improve performance on the related tasks.}

\subsection{\review{Multi-Task Models Applied to Natural Images}}
\review{\cite{bischke2019multi} proposed a cascaded multi-task loss to preserve boundary information from segmentation masks for segmenting building footprints and achieved state-of-the-art performance on an aerial image labeling task. \cite{he2017mask} extended Faster R-CNN~\citep{ren2015faster} by adding a new branch to predict the object mask along with a class label and a bounding box, and the proposed model was called Mask R-CNN. Mask R-CNN has been used extensively for multi-task segmentation models for a wide range of application areas~\citep{matterport_maskrcnn_2017}, such as adding sports fields to OpenStreetMap~\citep{jremillard_2018}, detection and segmentation for surgery robots~\citep{SUYEgit_2018}, understanding climate change patterns from aerial imagery of the Arctic~\citep{zhang2018deep}, converting satellite imagery to maps~\citep{crowdAIMappingChallengeBaseline2018}, detecting image forgeries~\citep{wang2019detection}, and segmenting tree canopy~\citep{Zhao2018ComparingUC}.}

\subsection{Multi-Task Models Applied to Medical Images}
\cite{chaichulee2017multi} extended the VGG16 architecture~\citep{simonyan2014very} to include a global average pooling layer for patient detection and a fully convolutional network for skin segmentation. The proposed model was evaluated on images from a clinical study conducted at a neonatal intensive care unit, and was robust to changes in lighting, skin tone, and pose.~\cite{He2019MultitaskLF} trained a U-Net~\citep{ronneberger2015u}-like encoder-decoder architecture to simultaneously segment thoracic organs from CT scans and perform global slice classification.~\cite{Ke2019AMU} trained a multi-task U-Net architecture to solve three tasks - separating wrongly connected objects, detecting class instances, and pixelwise labeling for each object, and evaluated it on a food microscopy image dataset. Other multi-task models have also been proposed for segmentation and classification for detecting manipulated faces in images and video~\citep{Nguyen2019MultitaskLF} and diagnosis of breast biopsy images~\citep{mehta2018net} and mammograms~\citep{le2019multitask}.

Mask R-CNN has also been used for segmentation tasks in medical image analysis such as automatically segmenting and tracking cell migration in phase-contrast microscopy~\citep{tsai2019usiigaci}, detecting and segmenting nuclei from histological and microscopic images~\citep{johnson2018adapting, vuola2019mask, wang2019multi, wang2019computational}, detecting and segmenting oral diseases~\citep{Anantharaman2018UtilizingMR}, segmenting neuropathic ulcers~\citep{gamage2019instance}, and labeling and segmenting ribs in chest X-rays~\citep{1908.08329}. Mask R-CNN has also been extended to work with 3D volumes and has been evaluated on lung nodule detection and segmentation from CT scans and breast lesion detection and categorization on diffusion MR images~\citep{jaeger2018retina, 1907.07676}.

\section{\review{Segmentation Evaluation Metrics and Datasets}}

\subsection{\review{Evaluation Metrics}}
\review{The quantitative evaluation of segmentation models can be performed using pixel-wise and overlap based measures. For binary segmentation, pixel-wise measures involve the construction of a confusion matrix to calculate the number of true positive (TP), true negative (TN), false positive (FP), and false negative (FN) pixels, and then calculate various metrics such as precision, recall (also known as sensitivity), specificity, and overall pixel-wise accuracy. They are defined as follows:}

\begin{equation}
    \textrm{Precision} = \frac{TP}{TP + FP},
\end{equation}
\begin{equation}
    \textrm{Recall or Sensitivity} = \frac{TP}{TP + FN},
\end{equation}
\begin{equation}
    \textrm{Specificity} = \frac{TN}{TN + FP}, \ \ \ \ \textrm{and,}
\end{equation}
\begin{equation}
    \textrm{Accuracy} = \frac{TP+TN}{TP + TN + FP + FN}.
\end{equation}

\review{Two popular overlap-based measures used to evaluate segmentation performance are the S{\o}rensen–Dice coefficient (also known as the Dice coefficient) and the Jaccard index (also known as the intersection over union or IoU). Given two sets $\mathcal{A}$ and $\mathcal{B}$, these metrics are defined as:}

\begin{equation}
    \textrm{Dice coefficient}, \textrm{Dice}(\mathcal{A},\mathcal{B}) = 2\ \frac{\left|\mathcal{A} \cap \mathcal{B}\right|}{\left|\mathcal{A}\right| + \left|\mathcal{B}\right|}, \ \ \ \ \textrm{and,}
\end{equation}
\begin{equation}
    \textrm{Jaccard index}, \textrm{Jaccard}(\mathcal{A},\mathcal{B}) = \frac{\left|\mathcal{A} \cap \mathcal{B}\right|}{\left|\mathcal{A} \cup \mathcal{B}\right|}.
\end{equation}

\review{For binary segmentation masks, these overlap-based measures can also be calculated from the confusion matrix as shown in Equations~\ref{dice-coef} and \ref{jac} respectively. The two measures are related by:}

\begin{equation}
    \textrm{Jaccard} = \frac{\textrm{Dice}}{2 - \textrm{Dice}}.
\end{equation}

\begin{figure}[!htbp]
     \centering
     \begin{subfigure}[b]{0.32\textwidth}
         \centering
         \includegraphics[width=0.4\textwidth]{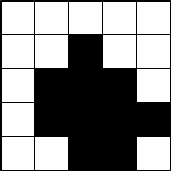}
         \caption{Ground truth binary mask}
         \label{subfig:SampleOverlap_GT}
     \end{subfigure}
     \hfill
     \begin{subfigure}[b]{0.32\textwidth}
         \centering
         \includegraphics[width=0.4\textwidth]{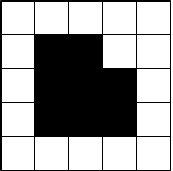}
         \caption{Predicted binary mask}
         \label{subfig:SampleOverlap_Pred}
     \end{subfigure}
     \hfill
     \begin{subfigure}[b]{0.32\textwidth}
         \centering
         \includegraphics[width=0.4\textwidth]{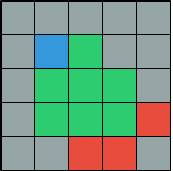}
         \caption{Overlap between the masks.}
         \label{subfig:SampleOverlap_Overlap}
     \end{subfigure}
        \caption{\review{A $5 \times 5$ overlap scenario with (a) the ground truth, (b) the predicted binary masks, and (c) the overlap. In (a) and (b), black and white pixels denote the foreground and the background respectively. In (c), green, grey, blue, and red pixels denote TP, TN, FP, and FN pixels respectively.}}
        \label{fig:SampleOverlap}
\end{figure}

\review{Figure~\ref{fig:SampleOverlap} contains a simple overlap scenario, with the ground truth and the predicted binary masks with a spatial resolution $5 \times 5$. Let black pixels denote the object to be segmented. The confusion matrix for this can be constructed as shown in Table~\ref{tab:SampleOverlap_Confusion}. Using the expressions above, we can calculate the metrics as $\textrm{precision} = \frac{7}{8}= 0.875$, $\textrm{recall} = \frac{7}{10}= 0.7$, $\textrm{specificity} = \frac{14}{15}= 0.9333$, $\textrm{pixel-wise accuracy} = \frac{21}{25}= 0.84$, $\textrm{Dice coefficient} = \frac{7}{9} = 0.7778$, and $\textrm{Jaccard index} = \frac{7}{11}= 0.6364$.}

\begin{table}[!h]
\centering
\caption{Confusion matrix for the overlap scenario shown in Figure~\ref{fig:SampleOverlap}.}
\label{tab:SampleOverlap_Confusion}
\begin{tabular}{cccccc}
\multicolumn{2}{c}{\multirow{2}{*}{}}                        & \multicolumn{2}{c}{\textbf{Ground Truth}}                  &  &  \\ 
\multicolumn{2}{c|}{}                                         & Background           & \multicolumn{1}{c}{Object} &  &  \\ \cline{2-6}
\multirow{2}{*}{\textbf{Prediction}} & \multicolumn{1}{c|}{Background} & 14                   & \multicolumn{1}{c}{3}      &  &  \\ 
                            & \multicolumn{1}{c|}{Object}     & 1                    & \multicolumn{1}{c}{7}      &  &  \\ \cline{2-6}
\end{tabular}
\end{table}

\subsection{\review{Semantic Segmentation Datasets for Natural Images}} \label{subsec:nat_datasets}

\review{Next, we briefly discuss the most popular and widely used datasets for the semantic segmentation of natural images. These datasets cover various categories of scenes, such as indoor and outdoor environments, common objects, urban street view as well as generic scenes. For a comprehensive review of the natural image datasets that segmentation models are usually benchmarked upon, we direct the interested readers to \cite{lateef2019survey}.}

\renewcommand{\arraystretch}{1.2}
\begin{table*}[htbp]
\centering
\caption{A summary of papers for semantic segmentation of natural images applied to PASCAL VOC 2012 dataset.}
\resizebox{\textwidth}{!}{%
\begin{tabular}{l|c|l|c}
\hline\hline
\textbf{Paper} & \textbf{\begin{tabular}[c]{@{}c@{}}Type of\\ Improvement\end{tabular}} & \textbf{Dataset(s) evaluated on} & \textbf{\begin{tabular}[c]{@{}c@{}}PASCAL VOC\\2012 mean IoU\end{tabular}}\\ 
\hline\hline

SegNet~(\cite{noh2015learning}) & Architecture    & \begin{tabular}[l]{@{}l@{}}PASCAL VOC, CamVid, SUN RGB-D\end{tabular} & 59.1\%  \\ \hline
FCN~(\cite{long2015fully}) & Architecture     & \begin{tabular}[l]{@{}l@{}}PASCAL VOC, NYUDv2, SIFT Flow\end{tabular} & 62.2\%    \\ \hline
~\cite{luc2016semantic}      & \begin{tabular}[c]{@{}c@{}}Adversarial\\Segmentation\end{tabular}                                                                                                                                      & \begin{tabular}[l]{@{}l@{}}PASCAL VOC, Stanford Background\end{tabular}          & 73.3\%\\ \hline
Lov{\'a}sz-Softmax Loss~\cite{matthew2018lovasz}           & Loss                                                                                                                                             & \begin{tabular}[c]{@{}c@{}}PASCAL VOC, Cityscapes\end{tabular}                 & 76.44\%\\ \hline
Large Kernel Matters~(\cite{peng2017large}) & Architecture    &   \begin{tabular}[l]{@{}l@{}}PASCAL VOC, Cityscapes\end{tabular}   &   82.2\%    \\  \hline
Deep Layer Cascade~(\cite{li2017not}) & Architecture    &   \begin{tabular}[l]{@{}l@{}}PASCAL VOC, Cityscapes\end{tabular}   &   82.7\%    \\  \hline
TuSimple~(\cite{wang2018understanding}) & Architecture    &   \begin{tabular}[l]{@{}l@{}}PASCAL VOC, KITTI Road Estimation\end{tabular}   &   83.1\%    \\  \hline
RefineNet~(\cite{lin2017refinenet}) & Architecture    &   \begin{tabular}[l]{@{}l@{}}PASCAL VOC, PASCAL Context, Person-Part,\\ NYUDv2, SUN RGB-D, Cityscapes, ADE20K\end{tabular}   &   84.2\%    \\  \hline
ResNet-38~(\cite{wu2019wider}) & Architecture    &   \begin{tabular}[l]{@{}l@{}}PASCAL VOC, PASCAL Context, Cityscapes\end{tabular}   &   84.9\%    \\  \hline
PSPNet~(\cite{zhao2017pyramid}) & Architecture    & PASCAL VOC, Cityscapes & 85.4\%                \\ \hline
Auto-DeepLab~(\cite{liu2019auto}) & \begin{tabular}[c]{@{}c@{}}Architecture\\Search\end{tabular}  & \begin{tabular}[l]{@{}l@{}}PASCAL VOC, ADE20K, Cityscapes\end{tabular} & 85.6\%            \\ \hline
IDW-CNN~(\cite{wang2017learning}) & Architecture    &   PASCAL VOC   &   86.3\%    \\  \hline
SDN+~(\cite{fu2019stacked}) & Architecture    &   \begin{tabular}[l]{@{}l@{}}PASCAL VOC, CamVid, Gatech\end{tabular}   &   86.6\%    \\  \hline
DIS~(\cite{luo2017deep}) & Architecture    &   PASCAL VOC   &   86.8\%    \\  \hline
DeepLabV3~(\cite{chen2017rethinking}) & Architecture  & PASCAL VOC & 86.9\%           \\ \hline
MSCI~(\cite{lin2018multi}) & Architecture     &   \begin{tabular}[l]{@{}l@{}}PASCAL VOC, PASCAL Context, NYUDv2,\\ SUN RGB-D\end{tabular}   & 88.0\%  \\    \hline
DeepLabV3+~(\cite{chen2018encoder}) & Architecture    & PASCAL VOC, Cityscapes & 89.0\%            \\ \hline
\hline
\end{tabular}
}
\label{tab:summary}
\end{table*}

\renewcommand{\arraystretch}{1.05}
\begin{table}[!ht]
\centering
\caption{\review{A summary of medical image segmentation papers along with their type of proposed improvement. \textbf{*} indicates the count at the highest level. For example, if a paper reports counts of patients, volumes, slices, etc., we report the count of patients.}}
\label{tab:summary_med}
\resizebox{\textwidth}{!}{%
\begin{tabular}{lll}
\hline\hline
\textbf{Paper}           & \textbf{Type of Improvement}                                             & \textbf{\begin{tabular}[c]{@{}l@{}}Training Data\\ {[}Modality (Site, Count*){]}\end{tabular}}                                                                                                                                                          \\ \hline\hline
\cite{ronneberger2015u}         & \begin{tabular}[c]{@{}l@{}}Architecture\\ Optimization\end{tabular}      & \begin{tabular}[c]{@{}l@{}}EM (Drosophilia, 30)\\ Microscopy (Cells, 35)\\ Microscopy (HeLa cells, 20)\end{tabular}                                                    \\ \hline
\cite{milletari2016v}           & \begin{tabular}[c]{@{}l@{}}Architecture\\ Optimization\end{tabular}      & MRI (Prostate, 50)                                                                                                                                                     \\ \hline
\cite{aicha_shapepriors}       & Optimization                                                             & Histology (Colon, 70)                                                                                                                                                  \\ \hline
\cite{yang2017fine}             & Sequenced Models                                                         & US (Prostate, 17)                                                                                                                                                      \\ \hline
\cite{salehi2017tversky}        & Optimization                                                             & MRI (Multiple sclerosis lesions. 15)                                                                                                                                   \\ \hline
\cite{chartsias2017adversarial} & Synthesis-based                                                          & CT (Heart, 20); MRI (Heart, 20)                                                                                                                                        \\ \hline
\cite{chaichulee2017multi}      & Multi-Task Models                                                        & RGB (Skin, 4603)                                                                                                                                                       \\ \hline
\cite{drozdzal2018learning}     & Architecture                                                             & \begin{tabular}[c]{@{}l@{}}EM (Drosophilia, 30)\\ CT (Liver, 77); MRI (Prostate, 50)\end{tabular}                                                                     \\ \hline
\cite{zhou2018unet++}           & Architecture                                                             & \begin{tabular}[c]{@{}l@{}}Microscopy (Cell nuclei, 670)\\ RGB (Colon polyp, 7379)\\ CT (Liver, 331)\\ CT (Lung nodule, 1012)\end{tabular}                             \\ \hline
\cite{oktay2018attention}       & Architecture                                                             & \begin{tabular}[c]{@{}l@{}}CT (Abdominal, 150)\\ CT (Pancreas, 82)\end{tabular}                                                                                        \\ \hline
\cite{xue2018segan}             & Architecture                                                             & MRI (Brain, 246)                                                                                                                                                       \\ \hline
\cite{zhang2018translating}     & Synthesis-based                                                          & CT (Heart, 4354); MRI (Heart, 142)                                                                                            \\ \hline
\cite{shin2018medical}          & Synthesis-based                                                          & MRI (Brain, 211)                                                                                                                                                       \\ \hline
\cite{johnson2018adapting}      & Multi-Task Models                                                        & Microscopy (Cell nuclei, 664)                                                                                                                                          \\ \hline
\cite{jaeger2018retina}         & Multi-Task Models                                                        & \begin{tabular}[c]{@{}l@{}}CT (Lung nodule, 1035)\\ MRI (Breast, 331)\end{tabular}                                                                                     \\ \hline
\cite{mehta2018net}             & Multi-Task Models                                                        & Histology (Breast, 428)                                                                                                                                                \\ \hline
\cite{huo2018adversarial}       & Synthesis-based                                                          & MRI (Spleen, 60); CT (Spleen, 19)                                                                                             \\ \hline
\cite{alom2019recurrent}        & Sequenced Models                                                         & \begin{tabular}[c]{@{}l@{}}Fundus (Retinal vessel, 20)\\ Fundus (Retinal vessel, 20)\\ Fundus (Retinal vessel, 28)\\ RGB (Skin, 1250); X-ray (Lung, 373)\end{tabular} \\ \hline
\cite{KERVADEC201988}           & \begin{tabular}[c]{@{}l@{}}Weakly Supervised\\ Optimization\end{tabular} & \begin{tabular}[c]{@{}l@{}}MRI (Heart, 75)\\ MRI (Vertebral body, 15)\\ MRI (Prostate, 40)\end{tabular}                                                                \\ \hline
\cite{Perone2019-eo}            & Weakly Supervised                                                        & MRI (Spinal cord, 40)                                                                                                                                                  \\ \hline\hline
\end{tabular}%
}
\end{table}

\begin{itemize}
    \item \review{\textbf{Pascal VOC datasets}: The PASCAL Visual Object Classes (VOC) Challenge~\citep{everingham2010pascal} was an annual challenge that ran from 2005 through 2012 and had annotations for several tasks such as classification, detection, and segmentation. The segmentation task was first introduced in the 2007 challenge and featured objects belonging to 20 classes. The last offering of the challenge, the PASCAL VOC 2012 challenge, contained segmentation annotations for 2,913 images across 20 object classes~\citep{everingham2015pascal}.}
    \item \review{\textbf{PASCAL Context}: The PASCAL Context dataset~\citep{mottaghi2014role} extended the PASCAL VOC 2010 Challenge dataset by providing pixel-wise annotations for the images, resulting in a much larger dataset with 19,740 annotated images and labels belonging to 540 categories.}
    \item \review{\textbf{Cityscapes}: The Cityscapes dataset~\citep{cordts2016cityscapes} contains annotated images of urban street scenes. The data was collected during daytime from 50 cities and exhibits variance in the season of the year and traffic conditions. Semantic, instance wise, and dense pixel-wise annotations are provided, with `fine' annotations for 5,000 images and `coarse' annotations for 20,000 images.}
    \item \review{\textbf{ADE20K}: The ADE20K dataset~\citep{zhou2017scene} contains 25,210 images from other existing datasets, e.g, the LabelMe~\citep{russell2008labelme}, the SUN~\citep{xiao2010sun}, and the Places~\citep{zhou2014learning} datasets. The images are annotated with labels belonging to 150 classes for ``scenes, objects, parts of objects, and in some cases even parts of parts''.}
    \item \review{\textbf{CamVid}: The Cambridge-driving Labeled Video Database (CamVid)~\citep{brostow2008segmentation,brostow2009semantic} contains 10 minutes of video captured at 30 frames per second from a driving automobile's perspective, along with pixel-wise semantic segmentation annotations for 701 frames and 32 object classes.}
\end{itemize}

Table~\ref{tab:summary} lists a summary of selected papers from this review, the nature of their proposed contributions, and the datasets that they were evaluated on. For the papers that evaluated their models on the PASCAL VOC 2012 dataset~\citep{pascal-voc-2012}, one of the most popular image semantic segmentation dataset for natural images, we also list their reported mean \review{IoU} scores. As can be seen in Table~\ref{tab:summary}, the focus has been mostly on architectural improvements. Comparing the first deep learning-based model (i.e., FCN~\citep{long2015fully}) to the state-of-the-art model (i.e., DeepLabV3+~\citep{chen2018encoder}) there is a large improvement (i.e. $\sim 27\%$, i.e., 62.2\% to 89.0\% ) in terms of mean IoU. The latter model leverages a more sophisticated decoder, dilated convolutions, and feature pyramid pooling.

\subsection{\review{Semantic Segmentation Datasets for Medical Images}}

\begin{figure}[!h]
     \centering
     \begin{subfigure}[b]{0.48\textwidth}
         \centering
         \includegraphics[width=\textwidth]{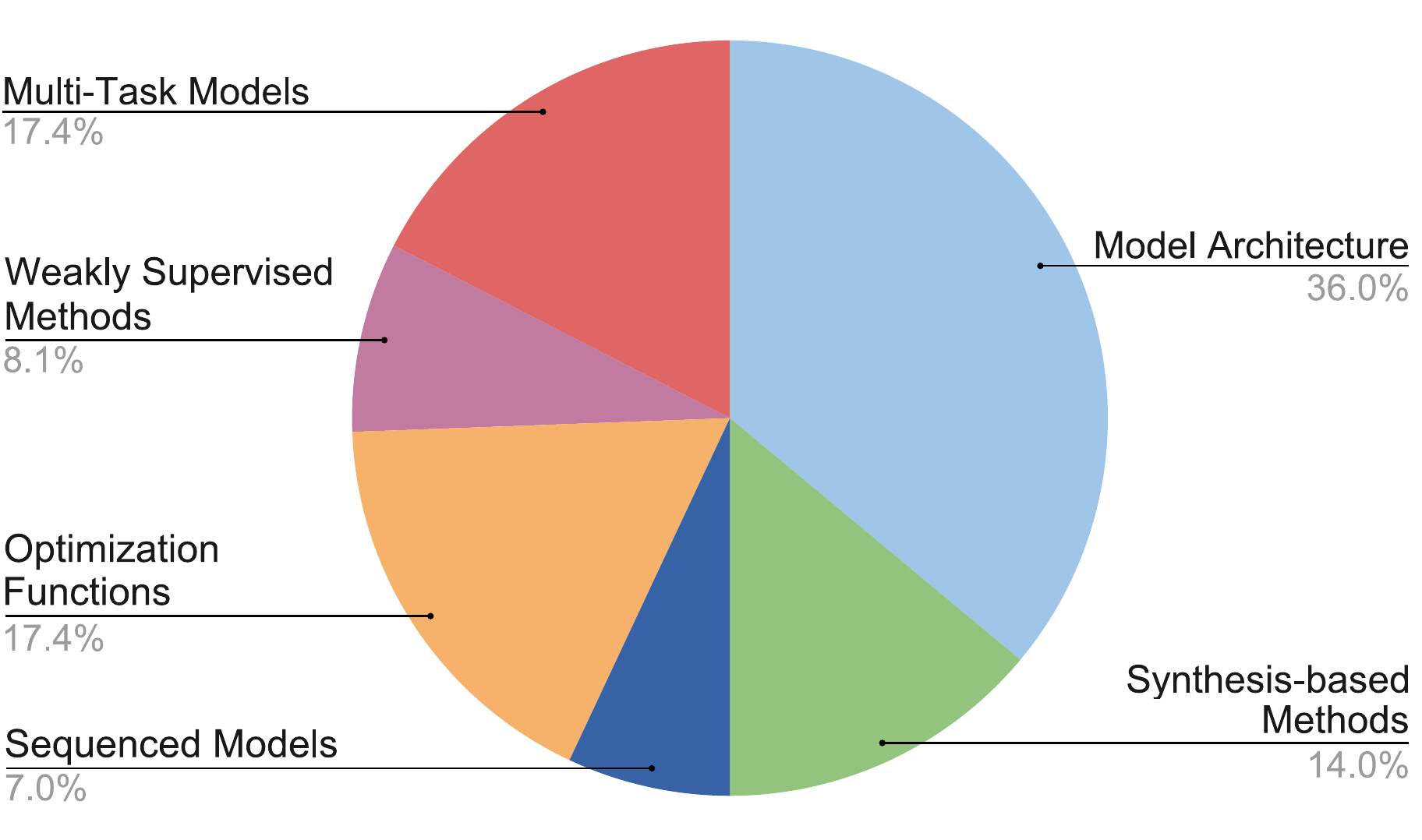}
         \caption{\review{The various categories of contributions.}}
     \end{subfigure}
     \quad
     \begin{subfigure}[b]{0.48\textwidth}
         \centering
         \includegraphics[width=\textwidth]{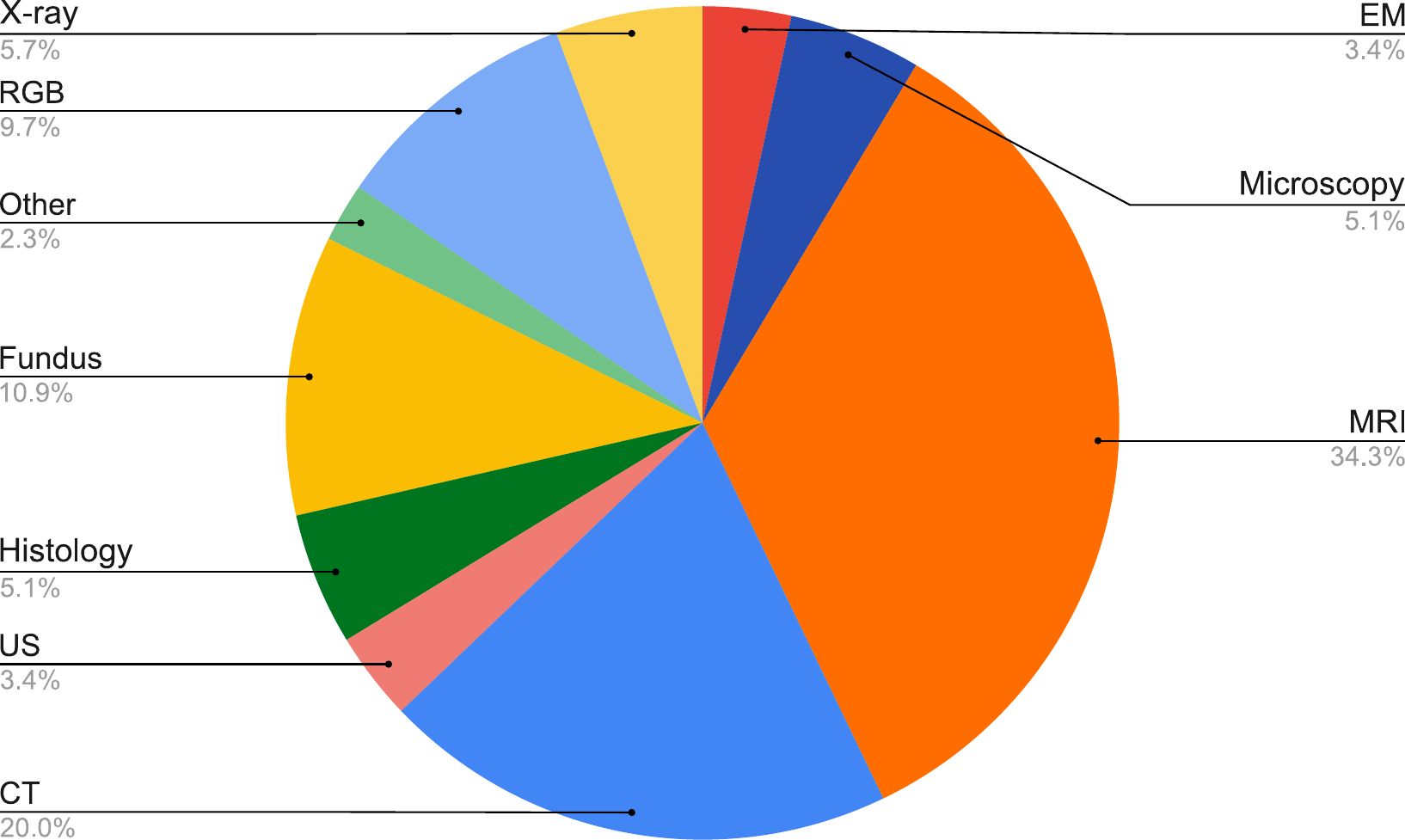}
         \caption{\review{The various medical imaging modalities.}}
     \end{subfigure}
     \quad
     \begin{subfigure}[b]{0.6\textwidth}
         \centering
         \includegraphics[width=\textwidth]{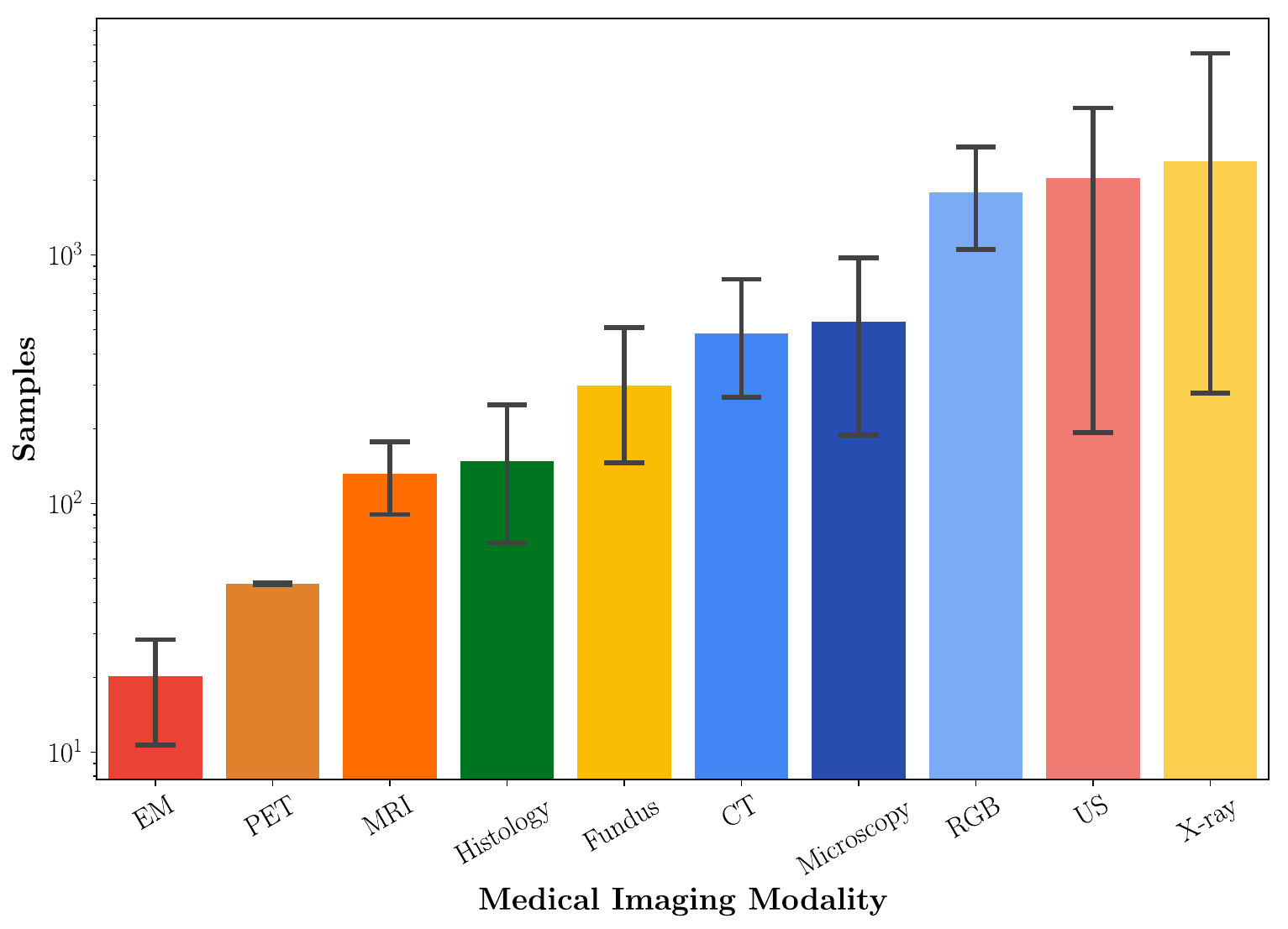}
         \caption{\review{The distribution of dataset sizes across multiple imaging modalities.}}
     \end{subfigure}
        \caption{\review{Analyzing the attributes of the medical image segmentation papers discussed in this review. The large number of medical imaging modalities (b) as well as the smaller average dataset sizes for medical image segmentation datasets (c) as compared to natural images (as discussed in Section~\ref{subsec:nat_datasets}) make it difficult to benchmark the performance of various approaches. In (b), PET (1.1\%), OCT (0.6\%), and topogram (0.6\%) make up the `Other' label.}}
        \label{fig:MedicalMeta}
\end{figure}

\begin{figure}[!h]
    \centering
    \includegraphics[scale=.4]{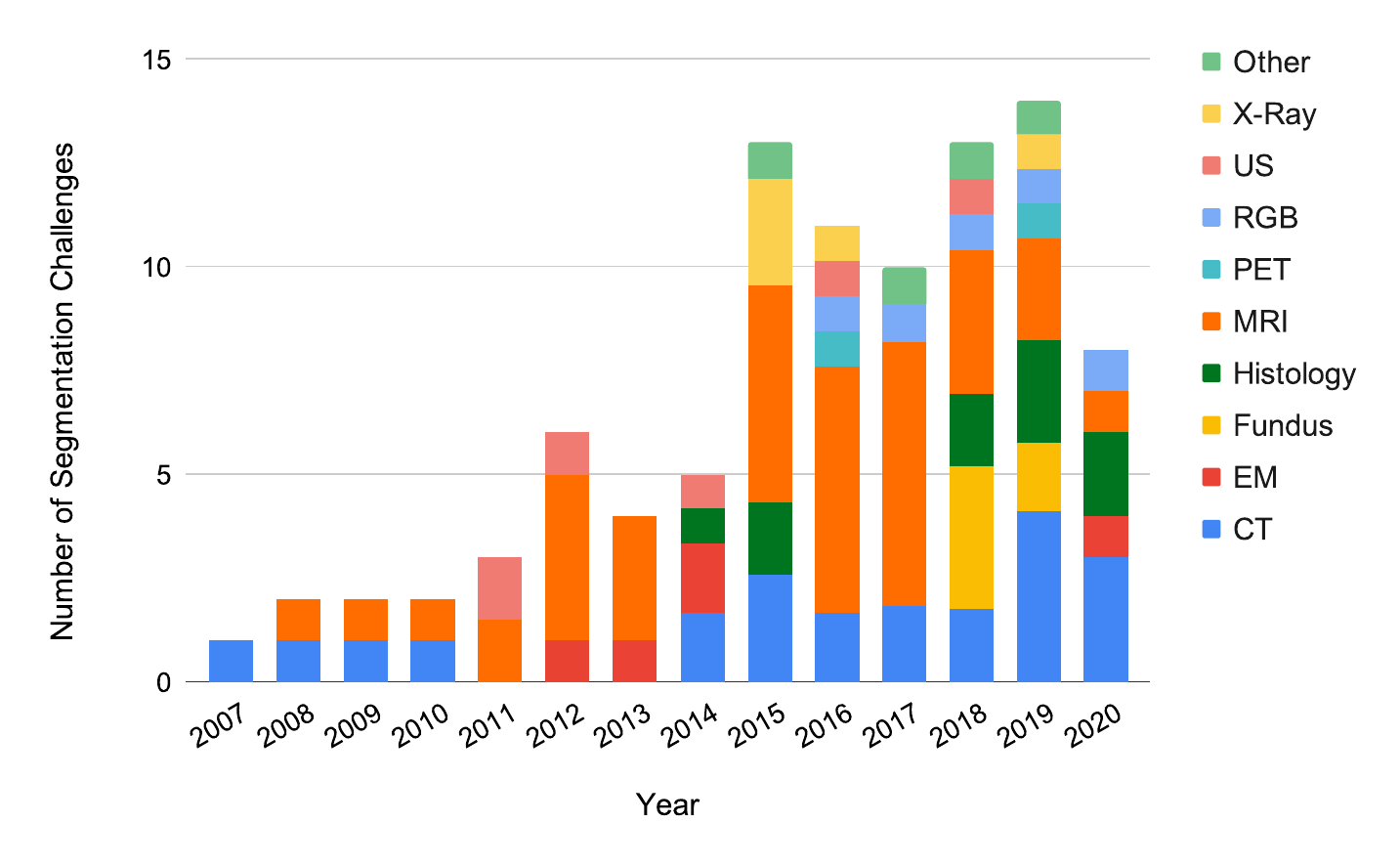}
    \caption {\review{The number of medical image segmentation challenges every year since 2007 listed on Grand Challenges~\citep{grand_challenge}}, along with a imaging modality-wise breakdown. Note that for many challenges, the data is multi-modal, and therefore the breakdown takes that into account.}
    \label{fig:grand_challenge}
\end{figure}

\review{In contrast to natural images, it is difficult to tabulate and summarize the performance of medical image segmentation methods because of the vast number of (a) medical imaging modalities and (b) medical image segmentation datasets. Figure~\ref{fig:MedicalMeta} presents a breakdown of the various attributes of the medical image segmentation papers surveyed in this review, color coded similar to Figure~\ref{fig:review}. As shown in Figure~\ref{fig:MedicalMeta} (b), the papers covered in this review use 13 medical imaging modalities. Figure~\ref{fig:MedicalMeta} (c) shows the distribution of the number of samples across datasets from multiple modalities. We observe that modalities which are expensive to acquire and annotate (such as electron microscopy (EM), PET, and MRI) have smaller dataset sizes than relative cheaper to acquire modalities such as RGB images (e.g., skin lesion images), ultrasound (US) and X-ray images. We also present a summary of the popular medical image segmentation papers in Table~\ref{tab:summary_med} and include the entire table in the Supplementary Material.}

\review{A similar observation can be made by looking at the medical image segmentation competitions. Grand Challenges in Biomedical Image Analysis~\citep{grand_challenge} provides a comprehensive but not exhaustive list of publicly available medical image segmentation challenges, and since 2007, there have been 94 segmentation challenges for medical images and volumes from as many as 12 imaging modalities. Figure~\ref{fig:grand_challenge} shows the number of these challenges for every year since 2007, and it can be seen that this number has been on the rise in the past few years.}

\section{Discussion and Future Directions}\label{disc}

In the following sections, we discuss in detail the potential future research directions for semantic segmentation of natural and medical images.

\subsection{Architectures}
Encoder-decoder networks with long and short skip connections are the winning architectures according to the state-of-the-art methods. Skip connections in deep networks have improved both segmentation and classification performance by facilitating the training of deeper network architectures and reducing the risks for vanishing gradients. They equip encoder-decoder-like networks with richer feature representations, but at the cost of higher memory usage, computation, and possibly resulting in transferring non-discriminative feature maps. Similar to~\citet{taghanaki2018select}, one future work direction is to optimize the amount of data is being transferred through skip connections. As for the cell level architectural design, our study shows that atrous convolutions with feature pyramid pooling modules are highly being used in the recent models. These approaches are somehow modifications of the classical convolution blocks. Similar to the radial basis function layers in~\citet{meyer2018deep} and \cite{taghanaki2019kernelized}, a future work focus can be designing new layers that capture a new aspect of data as opposed to convolutions or transform the convolution features into a new manifold. \review{Another useful research direction is using neural architecture search~\citep{zoph2016neural} to search for optimal deep neural network architectures for segmentation~\citep{liu2019auto,zhu2019neural,shaw2019squeezenas,weng2019neural}.}

\subsection{Sequenced Models}
For image segmentation, sequenced models can be used to segment temporal data such as videos. These models have also been applied to 3D medical datasets, however the advantage of processing volumetric data using 3D convolutions versus the processing the volume slice by slice using 2D sequenced models. Ideally, seeing the whole object of interest in a 3D volume might help to capture the geometrical information of the object, which might be missed in processing a 3D volume slice by slice. Therefore a future direction in this area can be through analysis of sequenced models versus volumetric convolution-based approaches.

\subsection{\review{Optimization} Functions}
In medical image segmentation works, researchers have converged toward using classical cross-entropy loss functions along with a second distance or overlap based functions. Incorporating domain/prior knowledge (such as coding the location of different organs explicitly in a deep model) is more sensible in the medical datasets. As shown in~\citet{taghanaki2019combo}, when only a distance-based or overlap-based loss function is used in a network, and the final layer applies sigmoid function, the risk of gradient vanishing increases. Although overlap based loss function are used in case of a class imbalance (small foregrounds), in Figure~\ref{ce_dice}, we show how using (\textit{only}) overlap based loss functions as the main term can be problematic for smooth optimization where they highly penalize a model under/over-segmenting a small foreground. However, the cross-entropy loss returns a reasonable score for the same cases. Besides using integrated cross-entropy based loss functions, future work can be exploring a single loss function that follows the behavior of the cross-entropy and at the same time, offers more features such capturing contour distance. This can be achieved by revisiting the current distance and overlap based loss functions. Another future path can be exploring auto loss function (or regularization term) search similar to the neural architecture search mentioned above. \review{Similarly, gradient based optimizations based on Sobolev~\citep{adams2003sobolev} gradients~\citep{czarnecki2017sobolev}, such as the works of~\cite{goceri2019diagnosis,goceri2020capsnet} are an interesting research direction.}

\subsection{Other Potential Directions}

\begin{itemize}

\item Going beyond pixel intensity-based scene understanding by incorporating prior knowledge\review{, which have been an active area of research for the past several decades~\citep{nosrati2016incorporating,xie2020survey}. Encoding prior knowledge in medical image analysis models is generally more possible as compared to natural images.} Currently, deep models receive matrices of intensity values, and usually, they are not forced to learn prior information. Without explicit reinforcement, the models might still learn object relations to some extent. However, it is difficult to interpret a learned strategy. 

\item Because of the large number of imaging modalities, the significant signal noise present in imaging modalities such as PET and ultrasound, and the limited amount of medical imaging data mainly because of high acquisition cost compounded by legal, ethical, and privacy issues, it is difficult to develop universal solutions that yield acceptable performances across various imaging modalities. Therefore, a proper research direction would be along the work of ~\cite{raghu2019transfusion} on image classification models, studying the risks of using non-medical pre-trained models for medical image segmentation. 
  
\item Creating large 2D and 3D publicly available medical benchmark datasets for semantic image segmentation such as the Medical Segmentation Decathlon~\citep{simpson2019large}. Medical imaging datasets are typically much smaller in size than natural image datasets~\citep{jin2020artificial}, and the curation of larger public datasets for medical imaging modalities will allow researchers to accurately compare proposed approaches and make incremental improvements for specific datasets and problems.

\item \review{A possible solution to address the paucity of sufficient annotated medical data is the development and use of physics based imaging simulators, the outputs of which can be used to train segmentation models and augment existing segmentation datasets. Several platforms~\citep{marion2011multi,glatard2013virtual} as well as simulators already exist for various imaging modalities such as SIMRI~\citep{benoitcattin2005simri} and POSSUM~\citep{drobnjak2006development,drobnjak2010simulating} for magnetic resonance imaging (MRI), PET-SORTEO~\citep{reilhac2005pet} and SimSET~\citep{harrison2012simset} for emission tomography, SINDBAD~\citep{tabary2007sindbad} for computed tomography (CT), and FIELD-II~\citep{jensen1992calculation,jensen1996field} and SIMUS~\citep{shahriari2018meshfree} for ultrasound imaging as well as anatomical regions of interest such as VascuSynth~\citep{hamarneh2010vascusynth} for vascular trees.}

\item \review{Medical images, both 2D and volumetric, have in general, larger file sizes than natural images, which inhibits the ability to load them entirely onto the memory for processing. As such, they need to be processed either as patches or sub-volumes, making it difficult for the segmentation models to capture spatial relationships in order to perform accurate segmentation. Therefore, an interesting and potentially very useful research direction would be coming up with architectures and training methods that can incorporate spatial relationships from large medical images and volumes in the models.}
  
\item Exploring reinforcement learning approaches similar to~\citet{song2018seednet} and~\citet{wang2018outline} for semantic (medical) image segmentation to mimic the way humans delineate objects of interest. Deep CNNs are successful in extracting features of different classes of objects, but they lose the local spatial information of where the borders of an object should be. Some researchers resort to traditional computer vision methods such as conditional random fields (CRFs) to overcome this problem, which however, add more computation time to the models.

\item Studying the causes for some models and datasets being prone to false positive and false negative predictions in the image segmentation context as found by ~\cite{matthew2018lovasz} and \cite{taghanaki2019combo}.
   
\item Exploring segmentation-free approaches~\citep{zhen2015towards,hussain2017segmentation,taghanaki2018segmentation,mukherjee2019fast,proenca2019segmentation}, i.e., bypassing the segmentation step according to the target problem. 

\item Weakly supervised segmentation using image-level labels versus a few images with segmentation annotations. Most new weakly supervised localization methods apply attention maps or region proposals in a multiple instance learning formulations. While attention maps can be noisy, leading to erroneously highlighted regions, it is not simple to decide on an optimal window or bag size for multiple instance learning approaches.

\item \review{While most deep segmentation models for medical image analysis rely on only clinical images for their predictions, there is often multi-modal patient data in the form of other imaging modalities as well as patient metadata that can provide valuable information, which most deep segmentation models do not use. Therefore, a valuable research direction for improving segmentation performance of medical images would be to develop models which are able to leverage multi-modal patient data.}

\item Modifying input instead of the model, loss function, and adding more train data.~\cite{drozdzal2018learning} showed that attaching a pre-processing module at the beginning of a segmentation network improves the network performance.~\cite{taghanaki2019improved} leveraged the gradients of a trained segmentation network with respect to the input to transfer it to a new space where the segmentation accuracy improves.

\item \review{Deep neural networks are trained using error backpropagation~\citep{Rumelhart1986} and gradient descent for optimizing the network weights. However, there have been many neural network optimization techniques which do not rely on backpropagation, such as credit assignment~\citep{bengio1994credit}, neuroevolution~\citep{stanley2002evolving}, difference target propagation~\citep{lee2015difference}, training with local error signals~\citep{pmlr-v97-nokland19a} and several other techniques~\citep{amit2019deep,bellec2019biologically,ma2019hsic}. Exploring these and similar other techniques to optimize deep neural networks for semantic segmentation would be another valuable research direction.}

\end{itemize}


%
%

\bibliographystyle{spbasic}      
\bibliography{bibliography.bib}   


\end{document}